\documentclass[twocolumn, 10pt]{article}

\pdfoutput=1

\usepackage{abstract}
\usepackage{amsmath}
\usepackage{amsthm}
\usepackage{bbm}
\usepackage{bm}
\usepackage{bibunits}
\usepackage{booktabs}
\usepackage{chngcntr}
\usepackage{etoolbox}
\usepackage[font=small,labelfont=sc]{caption}
\usepackage[T1]{fontenc}
\usepackage{fullpage}
\usepackage{graphicx}
\usepackage[colorlinks=true,
            linkcolor=blue,
            citecolor=blue,
            urlcolor=black]{hyperref}
\usepackage{lmodern}
\usepackage{longtable}
\usepackage{makecell}
\usepackage{multirow}
\usepackage[square]{natbib}
\usepackage{placeins}
\usepackage[detect-all]{siunitx}
\usepackage{subcaption}
\usepackage{tabularx}
\usepackage[table]{xcolor}
\usepackage{titling}

\defaultbibliography{rocs-features}
\defaultbibliographystyle{plainnat}

\robustify\bfseries
\definecolor{lightgray}{gray}{0.9}

\title{ROCS-Derived Features for Virtual Screening}

\author{
    \textbf{\normalsize Steven Kearnes} \\
    \normalsize Stanford University \\
    \texttt{\normalsize kearnes@stanford.edu} \and
    \textbf{\normalsize Vijay Pande} \\
    \normalsize Stanford University \\
    \texttt{\normalsize pande@stanford.edu}
}

\date{}  

\AtBeginDocument{}

\begin{document}
\begin{bibunit}
\maketitle

\begin{abstract}

Rapid overlay of chemical structures (ROCS) is a standard tool for the
calculation of 3D shape and chemical (``color'') similarity. ROCS uses
unweighted sums to combine many aspects of similarity, yielding parameter-free
models for virtual screening. In this report, we decompose the ROCS color force
field into \emph{color components} and \emph{color atom overlaps}, novel color
similarity features that can be weighted in a system-specific manner by machine
learning algorithms. In cross-validation experiments, these additional features
significantly improve virtual screening performance (ROC AUC scores) relative to
standard ROCS.

\end{abstract}


\section{Introduction}

Ligand-based virtual screening is based on the assumption that similar compounds
have similar biological activity~\citep{willett2009similarity}. Compound
similarity can be assessed in many ways, including comparisons of molecular
``fingerprints'' that encode structural features or molecular
properties~\citep{todeschini2009molecular} and measurements of shape, chemical,
and/or electrostatic similarity in three
dimensions~\citep{hawkins2007comparison, muchmore2006use,
ballester2007ultrafast}. Three-dimensional approaches such as rapid overlay of
chemical structures (ROCS)~\citep{hawkins2007comparison} are especially
interesting because of their potential to identify molecules that are similar
from the point of view of a target protein but dissimilar in underlying chemical
structure (``scaffold hopping'';~\citep{bohm2004scaffold}).

ROCS represents atoms as three-dimensional Gaussian
functions~\citep{grant1995gaussian, grant1996fast} and calculates similarity as
a function of volume overlaps between alignments of pre-generated molecular
conformers. Chemical (``color'') similarity is measured by overlaps between
dummy atoms marking interesting chemical functionalities: hydrogen bond donors
and acceptors, charged functional groups, rings, and hydrophobic groups. For
simplicity, the shape and color similarity scores are typically combined into a
single value that can be used to rank screening molecules against query
molecules with known activity. If more than one query molecule is available,
scores relative to each query can be combined using simple group fusion
methods such as max~\citep{chen2010combination}.

Machine learning methods offer powerful alternatives to combined similarity
scores and group fusion when additional experimental data is available for
training. By learning system-specific weights for the combination of similarity
features, these methods can avoid the loss of information that results from
combining these features arbitrarily (\emph{e.g.}, with an unweighted sum). For
example, \citet{sato2012application} showed that support vector machines (SVMs)
trained on ROCS similarity to a set of query molecules outperformed simple group
fusion models. Separating ROCS shape and color similarity scores and allowing
the model to weight them independently resulted in additional performance gains.

In this report, we extend the reductionism of \citet{sato2012application} by
decomposing ROCS color similarity scores into (1) separate components for each
color atom type (\emph{color components}) and (2) individual scores for color
atoms in query molecules (\emph{color atom overlaps}). We demonstrate
significant gains in virtual screening performance for machine learning models
trained on these features compared to standard ROCS and simpler partitioning of
shape and color similarity scores.

\section{Methods}
\label{sec:methods}

\subsection{ROCS features}
\label{sec:rocs_features}

All features were based on pairwise calculations by rapid overlay of chemical
structures (ROCS)~\citep{rocs}. ROCS measures the shape and chemical (``color'')
similarity of two compounds by calculating Tanimoto coefficients from aligned
overlap volumes:
\begin{equation}
\label{eq:tanimoto}
	\text{T}\left(A,B\right)=\frac{O_{AB}}{O_{AA}+O_{BB}-O_{AB}},
\end{equation}
where $O_{AB}$ is the aligned overlap volume between molecules $A$ and $B$.
Color similarity is calculated from overlaps between dummy atoms marking a
predefined set of pharmacophore features defined by the ROCS color force field.
The shape and color Tanimoto scores are often combined using an unweighted sum
or average to give a single similarity measure, TanimotoCombo. In typical ROCS
usage, one molecule is used as a reference or \emph{query} to search a screening
database or \emph{library} for similar compounds.

An alternative similarity measure, reference Tversky, emphasizes overlap with
the query molecule:
\begin{equation}
\label{eq:ref_tversky}
	\text{Tv}\left(A,B\right)=
	\frac{O_{AB}}{\alpha O_{AA}+ \left(1-\alpha\right)O_{BB}},
\end{equation}
where molecule A is the query and $\alpha$ varies the bias of the measurement
toward the query. In this report we used $\alpha=0.95$.

ROCS alignments, shape and color overlap volumes, and Tanimoto scores were
calculated using the \texttt{OEBestOverlay} object in the~\citet{oeshapetk}
(version 1.10.1). Overlays used the default Implicit Mills-Dean color force
field, \texttt{OEOverlapRadii\_All}, \texttt{OEOverlapMethod\_Analytic}, and the
following additional parameters: \texttt{color\_opt=True},
\texttt{use\_hydrogens=False}, \texttt{all\_color=True}.

\subsubsection{Color components}

The default color force field defines six color atom types---hydrogen bond
donors and acceptors, cations, anions, rings, and hydrophobes---and the volume
overlap terms use to calculate color Tanimoto scores are unweighted sums of the
overlap volumes for each color type. Since some pharmacophore types may be more
important than others in individual systems, we calculated separate similarity
scores for each color atom type. These scores are referred to as ROCS
\emph{color components}.

\subsubsection{Color atom overlaps}

In systems where query molecules have more than a single color atom of a given
type, ROCS color similarity scores do not contain information about the relative
locations of pharmacophore features. For example, a query molecule with two
hydrogen bond acceptors may give suboptimal color similarity scores for library
molecules if only one acceptor is important for activity. To avoid this problem
and allow models to distinguish between individual color features, we calculated
overlaps for individual color atoms in query molecules. These features are
referred to as ROCS \emph{color atom overlaps}.

\begin{figure*}[htb]
\centering
\includegraphics[clip,trim=0 160 30 0,width=\linewidth]{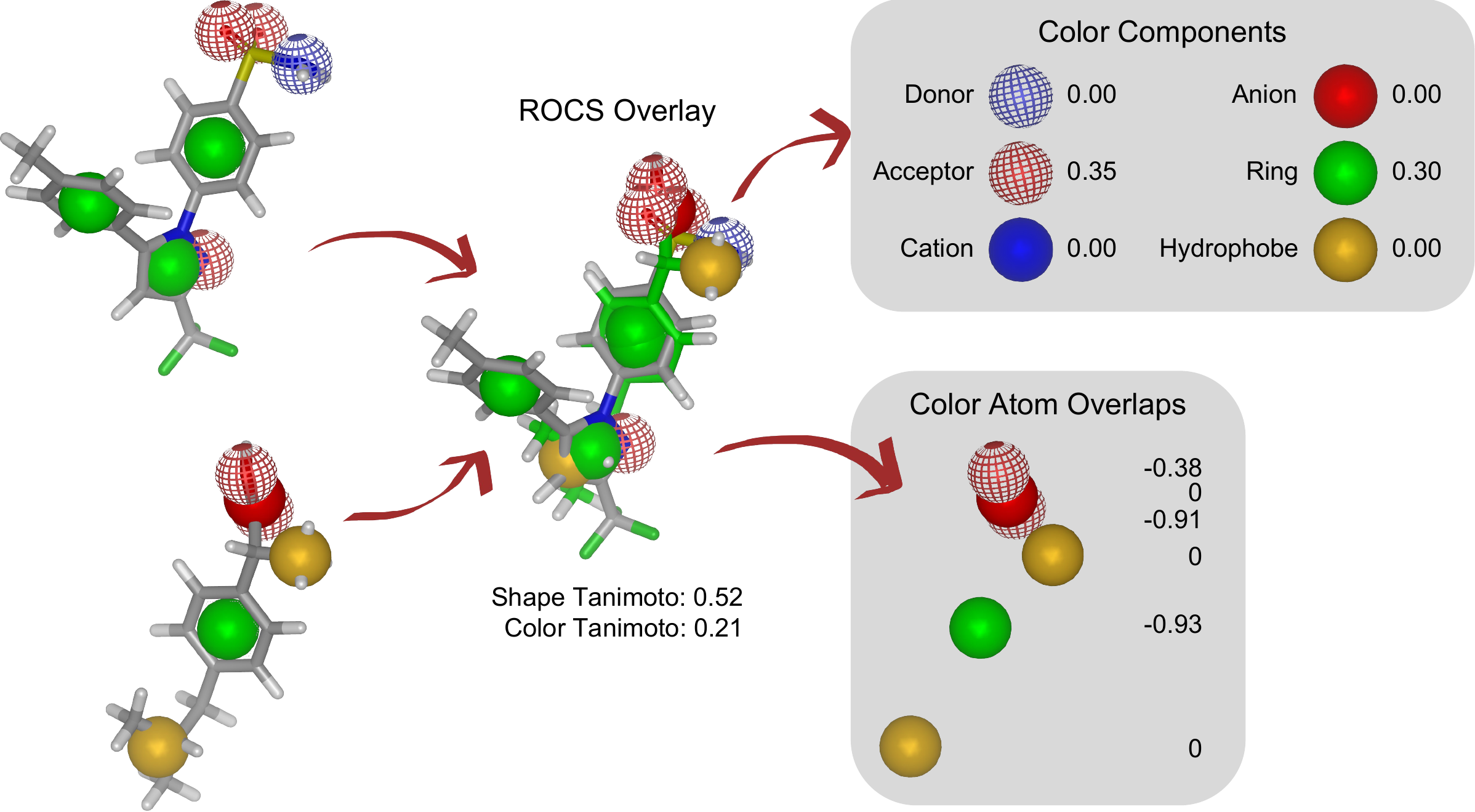}
\caption{Molecular descriptors based on the ROCS color force field. Color
components represent each color type independently. Color atom overlaps describe
similarity in terms of individual color atoms in the query molecule (bottom
left). Color component values are Tanimoto scores, but other similarity
metrics---such as reference Tversky---could be used as well. Note that color
atom overlap volumes are used without normalization, and that negative values
indicate favorable interactions.}
\label{fig:features}
\end{figure*}

The ROCS features used in this report, including color components and color atom
overlaps, are depicted in \figurename~\ref{fig:features}. We note that the
values used to calculate color components and color atom overlaps are available
internally to ROCS, but they are not directly accessible with the Shape Toolkit.

\subsection{Datasets}

We report results on three dataset collections with varying degrees of
difficulty. First, the enhanced directory of useful decoys (\mbox{DUD-E})
contains 102 datasets designed for validation of structure-based virtual
screening methods~\citep{mysinger2012directory}. Each dataset has an associated
protein structure and bound ligand. Second, the maximum unbiased validation
(MUV) collection contains 17 datasets designed for validation of ligand-based
virtual screening methods~\citep{rohrer2009maximum}. Each dataset contains a set
of maximally distinct actives individually embedded in a set of decoy molecules
to avoid analog bias and artificial enrichment. The third dataset collection was
derived from ChEMBL~\citep{gaulton2012chembl} for validation of ligand-based
methods~\citep{riniker2013open, riniker2013heterogeneous}. Each dataset (80 in
total) contains a set of diverse actives and shares a common set of decoys.

Up to 50 conformers for library molecules were generated with OpenEye
OMEGA~\citep{hawkins2010conformer, omega}. Query molecules were either used as
given (\mbox{DUD-E} crystal ligands) or with a single OMEGA conformer
(\mbox{DUD-E}, MUV, and ChEMBL).

By default, OMEGA does not generate conformers for molecules with unspecified
stereochemistry at chiral centers. This resulted in many compounds failing the
conformer generation step and consequently being excluded from our experiments.
Notably, 12 \mbox{DUD-E} crystal ligands failed OMEGA expansion and the corresponding
datasets were removed from the collection entirely, reducing this collection to
90 datasets. Additionally, about half of all ChEMBL compounds (actives and
decoys) failed conformational expansion due to unspecified stereochemistry.

The datasets in our collection are listed in Section~\ref{sec:datasets}, along
with counts of active and decoy molecules (not including OMEGA failures).

\subsection{Machine learning}

Standard ROCS is a parameter-free model that assigns the TanimotoCombo or other
combined similarity score relative to a query molecule as the positive class
probability. In situations where more than one query molecule is available,
group fusion methods such as max can be used to combine multiple similarity
scores into a single predicted value~\citep{chen2010combination}.
However, if more than one feature is used to describe
similarity or when more sophisticated combinations of multi-query similarities
are desired, machine learning or other statistical approaches can be used to
learn appropriate weights for each feature and tune performance for specific
systems.

Given a training set of $n$-dimensional feature vectors $\left\{\bm{x_1},
\bm{x_2}, \dots, \bm{x_m}\right\}$ with corresponding class labels $\left\{y_1,
y_2, \dots, y_m\right\}\in\left\{0,1\right\}$, a binary classifier learns a
decision function that predicts the positive class probability $\hat{y}$ of an
unlabeled feature vector $\bm{\hat{x}}$. The feature vectors are representations
of the input examples, encoding information that the classifier will attempt to
correlate to the output labels. Here, we use ROCS similarity scores and other
values derived from ROCS as features describing the relationships between query
and library molecules. For example, to learn system-specific weights for
combining shape and color Tanimotos relative to a single query, we would
construct feature vectors containing two elements corresponding to the separated
shape and color Tanimoto scores.

In this work, we trained three different types of binary classifiers: logistic
regression (LR), random forest (RF), and support vector machine (SVM). Logistic
regression is a simple linear classifier: a weight is assigned to each feature
and the output of the model is a biased linear combination of the input features
which is then nonlinearly scaled to the range $[0,1]$. Random forest is an
ensemble method which averages the output from several decision trees trained on
subsets of the input data~\citep{svetnik2003random}. Support vector machines are
maximum-margin classifiers that can model nonlinear relationships with the
appropriate choice of kernel function.

Classifiers were trained using scikit-learn 0.17.1~\citep{pedregosa2011scikit}.
Model hyperparameters ($C$ for LR and SVM) were tuned by stratified
cross-validation on training data. RF models used 100 trees and SVM models used
the RBF kernel. All models used the \texttt{`balanced'} class weighting
strategy. We also increased the maximum number of iterations for LR models to
\num{10000}.

\subsection{Model training and evaluation}
\label{sec:model_evaluation}

Models were trained using features calculated with respect to a single query
molecule using 5-fold stratified cross-validation and evaluated using metrics
calculated from the receiver operating characteristic (ROC)
curve~\citep{fawcett2006introduction}. The area under the ROC curve (AUC) is a
global measure of classification performance. ROC enrichment ($E_x$) measures
performance early in the ranked list of library molecules, calculated as the
ratio of the true positive rate (TPR) and the false positive rate (FPR) at a
specific FPR~\citep{jain2008recommendations}. We calculated ROC enrichment at
four FPR values: 0.005, 0.01, 0.02, and 0.05. The TPR corresponding to each FPR
was estimated by interpolation of the ROC curve generated by the
\texttt{roc\_curve} method in scikit-learn~\citep{pedregosa2011scikit}.

For each metric, we calculated mean test set values for each dataset across all
cross-validation folds. Per-dataset 5-fold mean metrics were further summarized
as medians within dataset collections (\mbox{DUD-E}, MUV, or ChEMBL) and
differences between methods are reported as median per-dataset $\Delta$AUC or
$\Delta E_x$ (here, $\Delta$ indicates a difference between 5-fold mean values).
Additionally, we report 95\% Wilson score intervals for the sign test statistic.
The sign test is a non-parametric statistical test that measures the fraction of
per-dataset differences that are greater than zero, and the confidence interval
is a measure of the expected consistency of an observed difference between two
methods. To calculate these intervals, we used the \texttt{proportion\_confint}
method in statsmodels~\citep{seabold2010statsmodels} with \texttt{alpha=0.05}
and \texttt{method=`wilson'}, ignoring any per-dataset differences that were
exactly zero.

For the \mbox{DUD-E} datasets, we trained models with the provided crystal
ligand as the query, using either the crystal coordinates or a single conformer
generated with OMEGA, resulting in 180 models total (90 datasets times two query
conformations). For each MUV and ChEMBL dataset, we trained 5-fold
cross-validation models specific to each active compound. For example, a MUV
dataset with 30 actives $\left\{a_1, a_2, \dots a_{30}\right\}$ resulted in 150
trained models corresponding to 30 rounds of 5-fold cross-validation, where the
features for round $i$ were specific to $a_i$ (which was removed from the
dataset before training).

When calculating median 5-fold mean AUC or ROC enrichment values and sign test
confidence intervals, the $N_i$ 5-fold models for each MUV and ChEMBL dataset
were treated as independent models rather than averaging across all queries.
This strategy allowed for more direct comparisons between models trained on the
same features and resulted in tighter confidence intervals than might be
expected (since the number of observations is greater than the number of
datasets). In total, we trained 378 5-fold models for MUV (17 datasets) and 4082
5-fold models for ChEMBL (80 datasets).

\section{Results}

\subsection{Proof of concept: \mbox{DUD-E}}

To assess the utility of \emph{color components} and \emph{color atom overlaps}
features (see Section~\ref{sec:rocs_features}), we trained models on various
combinations of input features using 5-fold cross-validation on the \mbox{DUD-E}
datasets. All models used shape Tanimoto (ST) along with some variant of color
similarity: color Tanimoto (CT), color component Tanimoto scores (CCT), and/or
color atom overlaps (CAO). (Abbreviations for features are consolidated in
\tablename~\ref{table:abbreviations}.) Each \mbox{DUD-E} dataset has a crystal
ligand that was used as the query molecule. As a comparative baseline, ROCS
TanimotoCombo scores were used to rank test set molecules by similarity to the
query; standard ROCS can be thought of as a model which assigns equal weight to
the shape and color Tanimoto scores. Additionally, we trained models using a
simple separation of shape and color similarity scores.

\begin{table}[tbp]
    \caption{Feature abbreviations.}
    \label{table:abbreviations}
    \centering
    \small
    \rowcolors{2}{}{lightgray}
    \begin{tabular}{ l l }
    \toprule
    Code & Description \\
    \midrule
    ST & Shape Tanimoto \\
    STv & Shape Tversky \\
    CT & Color Tanimoto \\
    CTv & Color Tversky \\
    CCT & Color components (Tanimoto scores) \\
    CCTv & Color components (Tversky scores) \\
    CAO & Query molecule color atom overlaps \\
    \bottomrule
    \end{tabular}
\end{table}

\tablename~\ref{table:dude} shows median 5-fold mean AUC scores for \mbox{DUD-E}
dataset models. For each model, we also report a two-sided 95\% confidence
interval around the mean difference in AUC relative to the ROCS baseline. ROC
enrichment scores for these models are reported in
Section~\ref{sec:roc_enrichment}. (Note that our analysis in this
report is based only on AUC scores.) In agreement with
\citet{sato2012application},
the ST-CT model achieved consistent improvements over ROCS by learning
target-specific weights for the combination of these features. Replacing the
color Tanimoto with color component scores (ST-CCT) gave an additional boost in
performance, and using color atom overlaps (ST-CAO) yielded even greater
improvement. Using color components and color atom overlaps in combination
(ST-CCT-CAO) gave additional improvements for median AUC and/or $\Delta$AUC
values, although these results were comparable to ST-CAO models (and were not
always more consistent, as measured by sign test confidence intervals).

It is common practice to scale input features in order to improve model training
and convergence. The results in \tablename~\ref{table:dude} were produced
without any feature scaling, but we experimented with two feature
transformations: scaling by maximum absolute value and ``standard'' scaling by
mean subtraction and division by the standard deviation (see
Section~\ref{sec:scaling}). Results for \mbox{DUD-E} crystal query models using
these feature scaling strategies are reported in
\tablename~\ref{table:max_abs_scaling} and
\tablename~\ref{table:standard_scaling}, respectively. Model performance was
relatively insensitive to feature scaling, and our subsequent analysis is based
on models trained without any feature transformations.

\begin{table*}[tbp]
    \caption{Model performance on \mbox{DUD-E} datasets. We report median AUC,
    median per-dataset $\Delta$AUC, and a sign test 95\% confidence interval
    (see Section~\ref{sec:model_evaluation}) for models that used either the
    crystal ligand or a generated conformer as the query. Bold values indicate
    statistically significant confidence intervals that do not include 0.5.}
    \label{table:dude}
    \centering
    \small
    \rowcolors{6}{}{lightgray}
    \sisetup{detect-weight=true,detect-inline-weight=math}
    \begin{tabular}{ l l S S c S S c }
    \toprule
     & & \multicolumn{3}{c}{Crystal Query Conformer} &
         \multicolumn{3}{c}{Generated Query Conformer} \\
    \cmidrule(lr){3-5} \cmidrule(lr){6-8}
    Model & Features &
    {\makecell{Median \\ AUC}} &
    {\makecell{Median \\ $\Delta$AUC}} &
    \makecell{Sign Test \\ 95\% CI} &
    {\makecell{Median \\ AUC}} &
    {\makecell{Median \\ $\Delta$AUC}} &
    \makecell{Sign Test \\ 95\% CI} \\
    \midrule
    \cellcolor{white} ROCS & TanimotoCombo & 0.697 &  &  & 0.633 &  &  \\
    \midrule
    \cellcolor{white} & ST-CT & 0.729 & \bfseries 0.023 & \bfseries (0.72, 0.88) & 0.698 & \bfseries 0.031 & \bfseries (0.72, 0.88) \\
    \cellcolor{white} & ST-CCT & 0.751 & \bfseries 0.056 & \bfseries (0.77, 0.91) & 0.730 & \bfseries 0.074 & \bfseries (0.79, 0.93) \\
    \cellcolor{white} & ST-CAO & 0.813 & \bfseries 0.123 & \bfseries (0.92, 0.99) & 0.813 & \bfseries 0.145 & \bfseries (0.92, 0.99) \\
    \multirow{-4}{*}{\cellcolor{white} LR} & ST-CCT-CAO & 0.823 & \bfseries 0.129 & \bfseries (0.91, 0.99) & 0.825 & \bfseries 0.158 & \bfseries (0.92, 0.99) \\
    \midrule
    \cellcolor{white} & ST-CT & 0.681 & 0.009 & (0.44, 0.64) & 0.678 & \bfseries 0.031 & \bfseries (0.52, 0.72) \\
    \cellcolor{white} & ST-CCT & 0.811 & \bfseries 0.124 & \bfseries (0.89, 0.98) & 0.810 & \bfseries 0.154 & \bfseries (0.88, 0.98) \\
    \cellcolor{white} & ST-CAO & 0.893 & \bfseries 0.198 & \bfseries (0.94, 1.00) & 0.888 & \bfseries 0.231 & \bfseries (0.94, 1.00) \\
    \multirow{-4}{*}{\cellcolor{white} RF} & ST-CCT-CAO & 0.893 & \bfseries 0.211 & \bfseries (0.94, 1.00) & 0.890 & \bfseries 0.244 & \bfseries (0.92, 0.99) \\
    \midrule
    \cellcolor{white} & ST-CT & 0.754 & \bfseries 0.052 & \bfseries (0.82, 0.95) & 0.746 & \bfseries 0.066 & \bfseries (0.78, 0.92) \\
    \cellcolor{white} & ST-CCT & 0.789 & \bfseries 0.085 & \bfseries (0.82, 0.95) & 0.778 & \bfseries 0.110 & \bfseries (0.83, 0.95) \\
    \cellcolor{white} & ST-CAO & 0.864 & \bfseries 0.167 & \bfseries (0.94, 1.00) & 0.856 & \bfseries 0.207 & \bfseries (0.92, 0.99) \\
    \multirow{-4}{*}{\cellcolor{white} SVM} & ST-CCT-CAO & 0.869 & \bfseries 0.171 & \bfseries (0.94, 1.00) & 0.867 & \bfseries 0.217 & \bfseries (0.92, 0.99) \\
    \bottomrule
    \end{tabular}
\end{table*}

We considered the possibility that these results were skewed by the use of a
crystal ligand as the query molecule. In most screening situations, a bioactive
conformation of the query is not known, and it is possible that color components
and color atom overlaps are more sensitive to the query conformation than
standard ROCS. Accordingly, we used OMEGA to generate conformers for
\mbox{DUD-E} crystal ligands and trained new models using generated conformers
as queries; results for these models are shown in \tablename~\ref{table:dude}.

As expected, standard ROCS performance decreased relative to models trained
using crystal query conformations (since generated conformers are not guaranteed
to represent bioactive conformations). However, separating shape and color
similarity or adding color components or color atom overlap features improved
performance in a manner consistent with crystal conformer queries. Notably, many
models achieved similar median AUC scores for both crystal and generated query
conformers, suggesting that these models were less sensitive to the query
conformation than standard ROCS.

\subsection{Additional benchmarks}

The \mbox{DUD-E} datasets were designed to avoid structural similarity between
active and inactive molecules in order to reduce the potential for false
negatives~\citep{mysinger2012directory}. Unfortunately, this aggravates issues
such as artificial enrichment and analog bias~\citep{rohrer2009maximum} and
limits their utility for validation of ligand-based
methods~\citep{irwin2008community}. To increase confidence in our results, we
trained models using ROCS-derived features on two additional dataset
collections: the maximum unbiased validation (MUV) datasets of
\citet{rohrer2009maximum} and a group of benchmarking datasets derived from
ChEMBL~\citep{riniker2013open, riniker2013heterogeneous}, both of which were
specifically designed for the validation of ligand-based methods. Since these
datasets are not associated with specific reference molecules (such as crystal
ligands), we trained multiple 5-fold models for each dataset using each active
molecule as a query (see Section~\ref{sec:model_evaluation}).

Performance metrics for models trained on MUV data are shown in
\tablename~\ref{table:muv}. MUV is known to be an especially challenging
benchmark, since each active molecule is explicitly separated from the others
and is embedded among hundreds of decoys with similar properties, so it was not
surprising that differences between MUV models were more variable and much
smaller than the differences observed with \mbox{DUD-E}. RF models
trained on MUV data were either no better or consistently worse than vanilla
ROCS. The only models that significantly outperformed the ROCS baseline were
trained on color atom overlap features, although sign test confidence intervals
for these models indicated that the benefit of including these features was not
as consistent for MUV as it was for \mbox{DUD-E}. ROC enrichment scores for
these models are reported in Section~\ref{sec:roc_enrichment}.

Results for models trained ChEMBL data are shown in
\tablename~\ref{table:chembl}. These datasets are more challenging than those in
\mbox{DUD-E}, yielding a substantially lower median ROC AUC for the ROCS
baseline (in fact, the ROCS median AUC for the ChEMBL datasets was lower
than for MUV). In contrast to the results for MUV, all ChEMBL machine learning models
saw consistent improvement over the ROCS baseline, with a ranking of feature
subsets similar to that observed for the \mbox{DUD-E} datasets. Notably, the
combination of color components and color atom overlaps (ST-CCT-CAO) resulted in
substantial improvements in median AUC relative to ST-CAO features for LR and
SVM models, although the sign test confidence intervals for these models were
similar. ROC enrichment
scores for these models are reported in Section~\ref{sec:roc_enrichment}.

\begin{table*}[tbp]
    \caption{Model performance on MUV datasets. We report median AUC, median
    per-dataset $\Delta$AUC, and a sign test 95\% confidence interval (see
    Section~\ref{sec:model_evaluation}). Several models were trained for each
    dataset, each using a different active molecule as the query. Bold values
    indicate statistically significant confidence intervals that do not include
    0.5.}
    \label{table:muv}
    \centering
    \small
    \rowcolors{2}{lightgray}{}
    \sisetup{detect-weight=true,detect-inline-weight=math}
    \begin{tabular}{ l l S S c }
    \toprule
    Model & Features &
    {\makecell{Median \\ AUC}} &
    {\makecell{Median \\ $\Delta$AUC}} &
    \makecell{Sign Test \\ 95\% CI} \\
    \midrule
    \cellcolor{white} ROCS & TanimotoCombo & 0.586 &  &  \\
    \midrule
    \cellcolor{white} & ST-CT & 0.603 & 0.001 & (0.46, 0.56) \\
    \cellcolor{white} & ST-CCT & 0.599 & 0.002 & (0.46, 0.56) \\
    \cellcolor{white} & ST-CAO & 0.615 & \bfseries 0.037 & \bfseries (0.58, 0.68) \\
    \multirow{-4}{*}{\cellcolor{white} LR} & ST-CCT-CAO & 0.632 & \bfseries 0.042 & \bfseries (0.60, 0.69) \\
    \midrule
    \cellcolor{white} & ST-CT & 0.517 & \bfseries \color{red} -0.059 & \bfseries \color{red} (0.25, 0.34) \\
    \cellcolor{white} & ST-CCT & 0.545 & \bfseries \color{red} -0.039 & \bfseries \color{red} (0.33, 0.43) \\
    \cellcolor{white} & ST-CAO & 0.562 & -0.005 & (0.43, 0.53) \\
    \multirow{-4}{*}{\cellcolor{white} RF} & ST-CCT-CAO & 0.566 & 0.000 & (0.45, 0.55) \\
    \midrule
    \cellcolor{white} & ST-CT & 0.597 & -0.002 & (0.44, 0.54) \\
    \cellcolor{white} & ST-CCT & 0.591 & 0.003 & (0.47, 0.57) \\
    \cellcolor{white} & ST-CAO & 0.609 & \bfseries 0.035 & \bfseries (0.56, 0.66) \\
    \multirow{-4}{*}{\cellcolor{white} SVM} & ST-CCT-CAO & 0.614 & \bfseries 0.040 & \bfseries (0.57, 0.67) \\
    \bottomrule
    \end{tabular}
\end{table*}

\begin{table*}[tbp]
    \caption{Model performance on ChEMBL datasets. We report median AUC, median
    per-dataset $\Delta$AUC, and a sign test 95\% confidence interval (see
    Section~\ref{sec:model_evaluation}). Several models were trained for each
    dataset, each using a different active molecule as the query. Bold values
    indicate statistically significant confidence intervals that do not include
    0.5.}
    \label{table:chembl}
    \centering
    \small
    \rowcolors{2}{lightgray}{}
    \sisetup{detect-weight=true,detect-inline-weight=math}
    \begin{tabular}{ l l S S c }
    \toprule
    Model & Features &
    {\makecell{Median \\ AUC}} &
    {\makecell{Median \\ $\Delta$AUC}} &
    \makecell{Sign Test \\ 95\% CI} \\
    \midrule
    \cellcolor{white} ROCS & TanimotoCombo & 0.579 &  &  \\
    \midrule
    \cellcolor{white} & ST-CT & 0.671 & \bfseries 0.037 & \bfseries (0.73, 0.76) \\
    \cellcolor{white} & ST-CCT & 0.716 & \bfseries 0.099 & \bfseries (0.83, 0.86) \\
    \cellcolor{white} & ST-CAO & 0.783 & \bfseries 0.180 & \bfseries (0.95, 0.96) \\
    \multirow{-4}{*}{\cellcolor{white} LR} & ST-CCT-CAO & 0.829 & \bfseries 0.223 & \bfseries (0.96, 0.97) \\
    \midrule
    \cellcolor{white} & ST-CT & 0.617 & \bfseries 0.032 & \bfseries (0.59, 0.62) \\
    \cellcolor{white} & ST-CCT & 0.744 & \bfseries 0.142 & \bfseries (0.85, 0.87) \\
    \cellcolor{white} & ST-CAO & 0.821 & \bfseries 0.213 & \bfseries (0.95, 0.97) \\
    \multirow{-4}{*}{\cellcolor{white} RF} & ST-CCT-CAO & 0.834 & \bfseries 0.224 & \bfseries (0.96, 0.97) \\
    \midrule
    \cellcolor{white} & ST-CT & 0.704 & \bfseries 0.088 & \bfseries (0.82, 0.85) \\
    \cellcolor{white} & ST-CCT & 0.752 & \bfseries 0.143 & \bfseries (0.88, 0.89) \\
    \cellcolor{white} & ST-CAO & 0.804 & \bfseries 0.200 & \bfseries (0.96, 0.97) \\
    \multirow{-4}{*}{\cellcolor{white} SVM} & ST-CCT-CAO & 0.835 & \bfseries 0.227 & \bfseries (0.96, 0.97) \\
    \bottomrule
    \end{tabular}
\end{table*}

\subsection{Model interpretation}

Color component and color atom overlap features provide fine detail on chemical
features that can be correlated with biological activity. Accordingly, trained
models can be interrogated for insights that are applicable to drug design. For
example, the learned weights for individual color atom overlaps in a linear
model contain information about the relative importance of those chemical
features for activity.

There are some technical details to keep in mind for the examples that follow.
Overlaps between color atoms are assigned negative feature values, such that
negative weights for color atoms indicate features that are correlated with
activity. Additionally, input features were not scaled during training or
prediction; the default color force field uses the same parameters for each
color atom interaction, such that learned weights for different color atoms
should be directly comparable. Shape similarity feature values ranged
from 0--1 (such that positive weights are correlated with activity); as such,
shape and color similarity features were not guaranteed to be on the same scale.

\begin{figure*}[htb]
  \centering
  \begin{subfigure}{0.49\linewidth}
    \includegraphics[clip,trim=190 190 280 220,width=\linewidth]{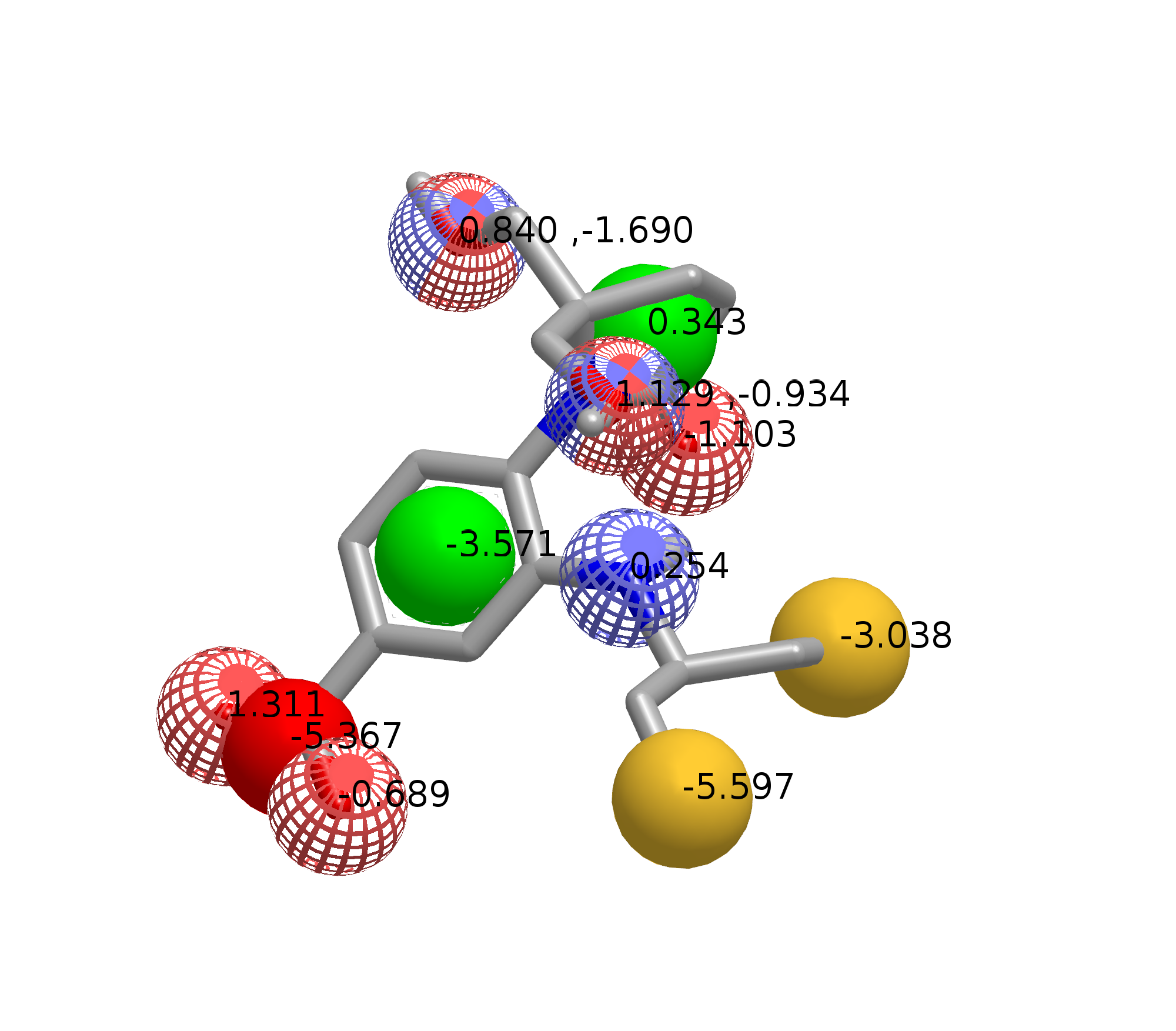}
    \caption{ST weight = \num{2.492}}
    \label{fig:nram}
  \end{subfigure}
  \begin{subfigure}{0.49\linewidth}
    \includegraphics[clip,trim=170 180 40 180,width=\linewidth]{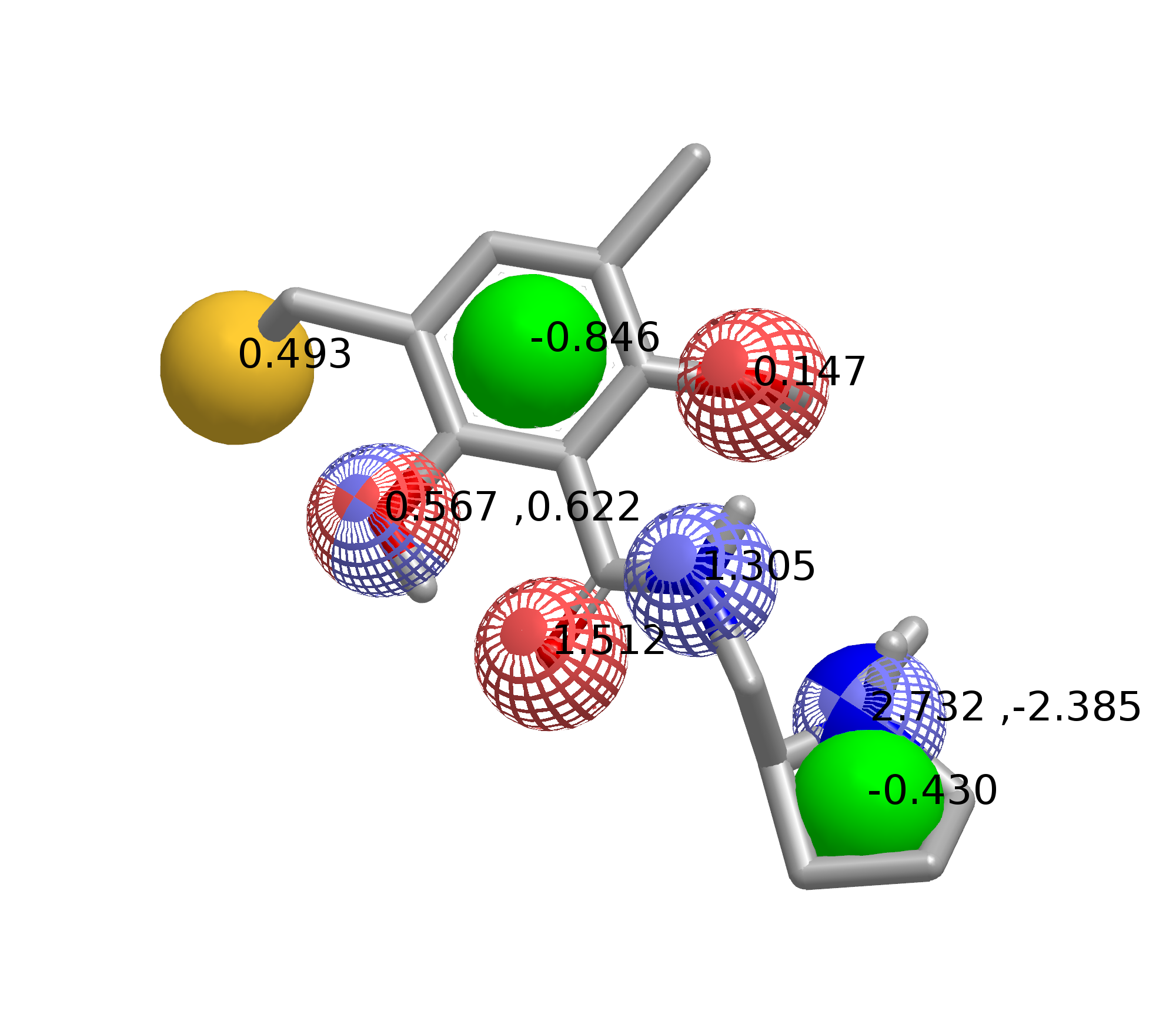}
    \caption{ST weight = \num{-2.270}}
    \label{fig:drd3}
  \end{subfigure}
  \caption{Learned color atom weights for the crystal ligands from the
  \mbox{DUD-E} (\subref{fig:nram}) nram (neuraminidase) and (\subref{fig:drd3})
  drd3 (dopamine receptor D3) datasets (refer to \figurename~\ref{fig:features}
  for a color atom type legend). In cases where two color atoms overlap, the
  listing order is (acceptor, donor) or (donor, cation). Negative weights on
  color atoms indicate features correlated with activity. Shape Tanimoto (ST)
  weights are also given for each model. Note that weights are from a model
  specific to a single cross-validation fold. Visualizations created with
  VIDA~\citep{vida} and the~\citet{oechemtk}.} \label{fig:model_weights}
\end{figure*}

\figurename~\ref{fig:nram} depicts the learned color atom weights from a
LR model trained on \mbox{ST-CAO} features for the \mbox{DUD-E}
nram (neuraminidase) dataset using the crystal ligand as the query molecule. In
this model, overlaps with the carboxylic acid anion, hydrophobic pentyl group,
and the aromatic ring are most correlated with activity. Interestingly, overlaps
with one of the carboxylic acid hydrogen bond acceptors are not favorable.
LR models trained on \mbox{ST-CAO} features achieved a 5-fold mean AUC of
\num{0.961}, compared to \num{0.862} for standard ROCS.


As another example, \figurename~\ref{fig:drd3} shows color atom weights from a
LR \mbox{ST-CAO} model for the \mbox{DUD-E} drd3 (dopamine receptor D3) dataset.
The overlapping donor and cation color atoms (lower right) are assigned positive
and negative weights, respectively. These weights suggest a potentially
counterintuitive result, that overlap with the query cation is important for
activity while the presence of a hydrogen bond donor at the same location is
correlated with inactivity. LR \mbox{ST-CAO} models substantially outperformed
vanilla ROCS on this dataset (\num{0.841} \emph{vs.} \num{0.313} 5-fold mean
AUC).

\subsection{Tversky features}

\citet{horvath2013not} demonstrated that reference Tversky---which biases
similarity scores to emphasize the features of the query molecule---can be a
more powerful metric than Tanimoto for virtual screening. Accordingly, we
repeated our analysis using reference Tversky variants of ROCS features; results
for these models are reported in Section~\ref{sec:tversky}. Baseline ROCS
performance using TverskyCombo was significantly higher for \mbox{DUD-E} and
ChEMBL datasets compared to TanimotoCombo, although the same general trends in
performance were observed for these datasets when training machine learning models with additional
features.
Notably, STv-CTv models tended to perform worse relative to the ROCS baseline
than their ST-CT/TanimotoCombo counterparts, with RF models performing
significantly worse than the TverskyCombo baseline in direct comparisons.
MUV models generally performed worse than or comparable to the
TverskyCombo baseline with the exception of LR models trained on color atom overlaps,
but in these cases the median differences were quite small and the associated sign test
confidence intervals were close to statistical insignificance.

\section{Discussion}

In this work, we described two new types of features derived from ROCS color
similarity scores: \emph{color components} that measure similarity for each
color type (\emph{e.g.} hydrogen bond donor, hydrophobe), and \emph{Color atom
overlaps} reporting overlap volumes for individual color atoms in query
molecules. Color atom overlaps provide spatial information to the model,
allowing overlaps with specific pharmacophore features to be considered in
predictions. We calculated both ROC AUC and ROC enrichment scores as measures of
model performance, but our analysis was based only on AUC scores.

Machine learning models trained using these features consistently outperformed
ROCS TanimotoCombo ranking in virtual screening experiments using datasets from
\mbox{DUD-E} and ChEMBL. Performance on MUV was less impressive, although modest
gains were observed for LR and SVM models trained with color atom overlap
features.

Additionally, we confirmed previous work showing the utility of reference
Tversky as a metric for virtual screening, and showed that models trained using
color components and color atom overlaps consistently outperformed ROCS
TverskyCombo baselines on the \mbox{DUD-E} and ChEMBL datasets.

We did not perform any experiments using more than one query molecule, but we
expect that color components and color atom overlaps will provide similar
benefits in multi-query situations.

Python code for generating color components and color atom overlaps features is
available online at \url{https://github.com/skearnes/color-features} and
requires a valid OpenEye Shape Toolkit license. The repository also includes
code for training and analysis of the models described in this report.

\section*{Acknowledgments}

We thank Paul Hawkins, Brian Cole, Anthony Nicholls, Brooke Husic, and Evan
Feinberg for helpful discussion. We also acknowledge use of the Stanford BioX3
cluster supported by NIH S10 Shared Instrumentation Grant 1S10RR02664701. S.K.
was supported by a Smith Stanford Graduate Fellowship. We also acknowledge
support from NIH 5U19AI109662-02.


\section*{Version information}

Submitted to the Journal of Computer-Aided Molecular Design.
Comments on arXiv versions:

\textbf{v2:} Fixed rounding of ROC enrichment confidence intervals and noted
that analysis is based only on ROC AUC scores.

\textbf{v3:} Added ``Model interpretation'' section, experiments with feature
scaling, and tables describing datasets. Some RF performance values changed
slightly due to model retraining. Also made updates throughout the text,
including a brief explanation of the method used to calculate ROC enrichment
and more thorough analysis in various Results sections.

\putbib
\end{bibunit}

\appendix
\counterwithin{figure}{section}
\counterwithin{table}{section}

\let\Oldsection\section
\renewcommand{\section}{\FloatBarrier\Oldsection}
\let\Oldsubsection\subsection
\renewcommand{\subsection}{\FloatBarrier\Oldsubsection}
\let\Oldsubsubsection\subsubsection
\renewcommand{\subsubsection}{\FloatBarrier\Oldsubsubsection}

\onecolumn
\begin{bibunit}


\section{Appendix: Datasets}
\label{sec:datasets}

The following tables provide information on the datasets used for building
models. The numbers of actives and decoys do not include compounds that failed
OMEGA expansion and may differ from their source publications. Note that 12/102
datasets from the original \mbox{DUD-E} publication were not used in this report
since their crystal ligands failed OMEGA expansion: aa2ar, andr, aofb, bace1,
braf, dyr, esr2, fkb1a, kif11, rxra, sahh, and urok.

\begin{table*}[htb]
    \caption{DUD-E datasets. Refer to~\citet{mysinger2012directory} for dataset
    descriptions and curation protocols. Note that 12/102 datasets from the
    original \mbox{DUD-E} publication were not used in this report since their
    crystal ligands failed OMEGA expansion.}
    \label{table:dude_datasets}
    \small
    \centering
    \rowcolors{1}{}{lightgray}
    \begin{tabular}[t]{ l S S S | }
    {Dataset} & {Actives} & {Decoys} & {\% Active} \\
    \midrule
abl1 & 165 & 9545 & 1.7 \\
ace & 136 & 16801 & 0.8 \\
aces & 373 & 20609 & 1.8 \\
ada & 51 & 5098 & 1.0 \\
ada17 & 394 & 34570 & 1.1 \\
adrb1 & 166 & 13434 & 1.2 \\
adrb2 & 151 & 12720 & 1.2 \\
akt1 & 244 & 13649 & 1.8 \\
akt2 & 106 & 6042 & 1.7 \\
aldr & 143 & 8958 & 1.6 \\
ampc & 43 & 2835 & 1.5 \\
cah2 & 414 & 29968 & 1.4 \\
casp3 & 109 & 10437 & 1.0 \\
cdk2 & 428 & 26499 & 1.6 \\
comt & 38 & 3840 & 1.0 \\
cp2c9 & 105 & 7089 & 1.5 \\
cp3a4 & 129 & 10709 & 1.2 \\
csf1r & 159 & 11059 & 1.4 \\
cxcr4 & 24 & 2720 & 0.9 \\
def & 68 & 5505 & 1.2 \\
dhi1 & 252 & 18719 & 1.3 \\
dpp4 & 452 & 36118 & 1.2 \\
drd3 & 381 & 28391 & 1.3 \\
egfr & 497 & 31707 & 1.5 \\
esr1 & 286 & 18303 & 1.5 \\
fa10 & 273 & 25387 & 1.1 \\
fa7 & 18 & 5877 & 0.3 \\
fabp4 & 45 & 2741 & 1.6 \\
fak1 & 97 & 5052 & 1.9 \\
fgfr1 & 134 & 7755 & 1.7 \\
fnta & 372 & 46853 & 0.8 \\
fpps & 72 & 8800 & 0.8 \\
gcr & 126 & 14570 & 0.9 \\
glcm & 29 & 3481 & 0.8 \\
gria2 & 133 & 11271 & 1.2 \\
grik1 & 90 & 6505 & 1.4 \\
hdac2 & 168 & 9834 & 1.7 \\
hdac8 & 157 & 9942 & 1.6 \\
hivint & 79 & 6591 & 1.2 \\
hivpr & 291 & 34714 & 0.8 \\
hivrt & 260 & 18476 & 1.4 \\
hmdh & 89 & 8726 & 1.0 \\
hs90a & 88 & 4668 & 1.9 \\
hxk4 & 80 & 4650 & 1.7 \\
igf1r & 138 & 8417 & 1.6 \\
    \end{tabular}
    \rowcolors{1}{}{lightgray}
    \begin{tabular}[t]{ | l S S S }
    {Dataset} & {Actives} & {Decoys} & {\% Active} \\
    \midrule
inha & 31 & 2256 & 1.4 \\
ital & 117 & 8140 & 1.4 \\
jak2 & 97 & 5942 & 1.6 \\
kit & 163 & 9449 & 1.7 \\
kith & 38 & 2777 & 1.3 \\
kpcb & 101 & 8110 & 1.2 \\
lck & 389 & 24717 & 1.5 \\
lkha4 & 153 & 7952 & 1.9 \\
mapk2 & 74 & 5969 & 1.2 \\
mcr & 34 & 5059 & 0.7 \\
met & 155 & 10090 & 1.5 \\
mk01 & 67 & 4436 & 1.5 \\
mk10 & 93 & 6122 & 1.5 \\
mk14 & 548 & 32665 & 1.6 \\
mmp13 & 398 & 35932 & 1.1 \\
mp2k1 & 93 & 7338 & 1.3 \\
nos1 & 51 & 6887 & 0.7 \\
nram & 68 & 6145 & 1.1 \\
pa2ga & 87 & 5127 & 1.7 \\
parp1 & 411 & 28524 & 1.4 \\
pde5a & 329 & 25738 & 1.3 \\
pgh1 & 165 & 10719 & 1.5 \\
pgh2 & 374 & 22997 & 1.6 \\
plk1 & 101 & 6318 & 1.6 \\
pnph & 75 & 6842 & 1.1 \\
ppara & 264 & 19287 & 1.4 \\
ppard & 201 & 12201 & 1.6 \\
pparg & 332 & 25133 & 1.3 \\
prgr & 177 & 15429 & 1.1 \\
ptn1 & 90 & 7216 & 1.2 \\
pur2 & 21 & 2680 & 0.8 \\
pygm & 44 & 3887 & 1.1 \\
pyrd & 109 & 6423 & 1.7 \\
reni & 71 & 6041 & 1.2 \\
rock1 & 64 & 5617 & 1.1 \\
src & 487 & 30101 & 1.6 \\
tgfr1 & 122 & 7967 & 1.5 \\
thb & 93 & 7249 & 1.3 \\
thrb & 209 & 24898 & 0.8 \\
try1 & 84 & 23602 & 0.4 \\
tryb1 & 111 & 7015 & 1.6 \\
tysy & 73 & 6568 & 1.1 \\
vgfr2 & 392 & 22056 & 1.7 \\
wee1 & 97 & 5868 & 1.6 \\
xiap & 88 & 4638 & 1.9 \\
    \end{tabular}
\end{table*}

\begin{table*}[htb]
    \caption{MUV datasets. Refer to~\citet{rohrer2009maximum} for dataset
    descriptions and curation protocols.}
    \label{table:muv_datasets}
    \small
    \centering
    \rowcolors{1}{}{lightgray}
    \begin{tabular}[t]{ l S S S }
    {Dataset} & {Actives} & {Decoys} & {\% Active} \\
    \midrule
aid466 & 23 & 11541 & 0.2 \\
aid548 & 26 & 11604 & 0.2 \\
aid600 & 23 & 10772 & 0.2 \\
aid644 & 24 & 11133 & 0.2 \\
aid652 & 18 & 11103 & 0.2 \\
aid689 & 13 & 9948 & 0.1 \\
aid692 & 24 & 10934 & 0.2 \\
aid712 & 18 & 10288 & 0.2 \\
aid713 & 24 & 11641 & 0.2 \\
aid733 & 21 & 10855 & 0.2 \\
aid737 & 26 & 11219 & 0.2 \\
aid810 & 19 & 11186 & 0.2 \\
aid832 & 24 & 11093 & 0.2 \\
aid846 & 28 & 11274 & 0.2 \\
aid852 & 27 & 10865 & 0.2 \\
aid858 & 20 & 11275 & 0.2 \\
aid859 & 20 & 10735 & 0.2 \\
    \end{tabular}
\end{table*}

\begin{table*}[htb]
    \caption{ChEMBL datasets. Refer to~\citet{riniker2013open,
    riniker2013heterogeneous} for dataset descriptions and curation protocols.
    Note that all ChEMBL datasets share the same set of decoys.}
    \label{table:chembl_datasets}
    \small
    \centering
    \rowcolors{1}{}{lightgray}
    \begin{tabular}[t]{ l S S S | }
    {Dataset} & {Actives} & {Decoys} & {\% Active} \\
    \midrule
chembl100126 & 81 & 4822 & 1.7 \\
chembl100166 & 34 & 4822 & 0.7 \\
chembl100579 & 57 & 4822 & 1.2 \\
chembl100 & 30 & 4822 & 0.6 \\
chembl10188 & 76 & 4822 & 1.6 \\
chembl10193 & 57 & 4822 & 1.2 \\
chembl10198 & 19 & 4822 & 0.4 \\
chembl10260 & 57 & 4822 & 1.2 \\
chembl10280 & 58 & 4822 & 1.2 \\
chembl10378 & 16 & 4822 & 0.3 \\
chembl10417 & 44 & 4822 & 0.9 \\
chembl10434 & 46 & 4822 & 0.9 \\
chembl10475 & 39 & 4822 & 0.8 \\
chembl10498 & 28 & 4822 & 0.6 \\
chembl104 & 47 & 4822 & 1.0 \\
chembl10579 & 66 & 4822 & 1.4 \\
chembl105 & 69 & 4822 & 1.4 \\
chembl10752 & 73 & 4822 & 1.5 \\
chembl10773 & 47 & 4822 & 1.0 \\
chembl107 & 51 & 4822 & 1.0 \\
chembl108 & 58 & 4822 & 1.2 \\
chembl10927 & 14 & 4822 & 0.3 \\
chembl10980 & 59 & 4822 & 1.2 \\
chembl11085 & 73 & 4822 & 1.5 \\
chembl11140 & 13 & 4822 & 0.3 \\
chembl11225 & 17 & 4822 & 0.4 \\
chembl11265 & 74 & 4822 & 1.5 \\
chembl11279 & 91 & 4822 & 1.9 \\
chembl11336 & 81 & 4822 & 1.7 \\
chembl11359 & 65 & 4822 & 1.3 \\
chembl11365 & 51 & 4822 & 1.0 \\
chembl11442 & 39 & 4822 & 0.8 \\
chembl11488 & 73 & 4822 & 1.5 \\
chembl11489 & 51 & 4822 & 1.0 \\
chembl114 & 76 & 4822 & 1.6 \\
chembl11534 & 32 & 4822 & 0.7 \\
chembl11536 & 28 & 4822 & 0.6 \\
chembl11575 & 27 & 4822 & 0.6 \\
chembl11631 & 42 & 4822 & 0.9 \\
chembl11682 & 46 & 4822 & 0.9 \\
    \end{tabular}
    \rowcolors{1}{}{lightgray}
    \begin{tabular}[t]{ | l S S S }
    {Dataset} & {Actives} & {Decoys} & {\% Active} \\
    \midrule
chembl116 & 56 & 4822 & 1.1 \\
chembl121 & 34 & 4822 & 0.7 \\
chembl12209 & 62 & 4822 & 1.3 \\
chembl12252 & 24 & 4822 & 0.5 \\
chembl12261 & 81 & 4822 & 1.7 \\
chembl12670 & 62 & 4822 & 1.3 \\
chembl12679 & 31 & 4822 & 0.6 \\
chembl126 & 64 & 4822 & 1.3 \\
chembl12840 & 78 & 4822 & 1.6 \\
chembl12911 & 53 & 4822 & 1.1 \\
chembl12952 & 69 & 4822 & 1.4 \\
chembl12968 & 34 & 4822 & 0.7 \\
chembl13001 & 31 & 4822 & 0.6 \\
chembl130 & 45 & 4822 & 0.9 \\
chembl134 & 54 & 4822 & 1.1 \\
chembl15 & 53 & 4822 & 1.1 \\
chembl165 & 37 & 4822 & 0.8 \\
chembl17045 & 39 & 4822 & 0.8 \\
chembl18061 & 42 & 4822 & 0.9 \\
chembl19905 & 49 & 4822 & 1.0 \\
chembl20014 & 78 & 4822 & 1.6 \\
chembl20174 & 72 & 4822 & 1.5 \\
chembl219 & 49 & 4822 & 1.0 \\
chembl234 & 76 & 4822 & 1.6 \\
chembl237 & 55 & 4822 & 1.1 \\
chembl259 & 48 & 4822 & 1.0 \\
chembl25 & 12 & 4822 & 0.2 \\
chembl276 & 80 & 4822 & 1.6 \\
chembl28 & 66 & 4822 & 1.4 \\
chembl36 & 35 & 4822 & 0.7 \\
chembl43 & 36 & 4822 & 0.7 \\
chembl51 & 51 & 4822 & 1.0 \\
chembl52 & 55 & 4822 & 1.1 \\
chembl61 & 55 & 4822 & 1.1 \\
chembl65 & 50 & 4822 & 1.0 \\
chembl72 & 48 & 4822 & 1.0 \\
chembl87 & 53 & 4822 & 1.1 \\
chembl8 & 53 & 4822 & 1.1 \\
chembl90 & 63 & 4822 & 1.3 \\
chembl93 & 44 & 4822 & 0.9 \\
    \end{tabular}
\end{table*}

\section{Appendix: Feature scaling}
\label{sec:scaling}

The following tables report results on \mbox{DUD-E} datasets using crystal
ligand queries for different feature scaling strategies.
\tablename~\ref{table:max_abs_scaling} gives results for features scaled by
maximum absolute value with the scikit-learn~\citep{pedregosa2011scikit}
\texttt{MaxAbsScaler} class. \tablename~\ref{table:standard_scaling} gives
results for features scaled by mean subtraction and division by the standard
deviation using the scikit-learn \texttt{StandardScaler} class. For both scaling
strategies, scaling parameters were learned from training data and then applied
to the dataset as a whole (within each cross-validation fold).

\begin{table*}[tbp]
    \caption{Model performance on \mbox{DUD-E} datasets using crystal conformer
    queries with input features scaled by maximum absolute value. We report
    median AUC, median per-dataset $\Delta$AUC, and a sign test 95\% confidence
    interval (see Section~\ref{sec:model_evaluation}). Bold values indicate
    statistically significant confidence intervals that do not include 0.5.}
    \label{table:max_abs_scaling}
    \centering
    \small
    \rowcolors{2}{lightgray}{}
    \sisetup{detect-weight=true,detect-inline-weight=math}
    \begin{tabular}{ l l S S c }
    \toprule
    Model & Features &
    {\makecell{Median \\ AUC}} &
    {\makecell{Median \\ $\Delta$AUC}} &
    \makecell{Sign Test \\ 95\% CI} \\
    \midrule
    \cellcolor{white} ROCS & TanimotoCombo & 0.697 &  &  \\
    \midrule
    \cellcolor{white} & ST-CT & 0.730 & \bfseries 0.023 & \bfseries (0.72, 0.88) \\
    \cellcolor{white} & ST-CCT & 0.756 & \bfseries 0.057 & \bfseries (0.78, 0.92) \\
    \cellcolor{white} & ST-CAO & 0.813 & \bfseries 0.118 & \bfseries (0.92, 0.99) \\
    \multirow{-4}{*}{\cellcolor{white} LR} & ST-CCT-CAO & 0.826 & \bfseries 0.126 & \bfseries (0.92, 0.99) \\
    \midrule
    \cellcolor{white} & ST-CT & 0.677 & 0.005 & (0.43, 0.63) \\
    \cellcolor{white} & ST-CCT & 0.809 & \bfseries 0.125 & \bfseries (0.88, 0.98) \\
    \cellcolor{white} & ST-CAO & 0.894 & \bfseries 0.201 & \bfseries (0.94, 1.00) \\
    \multirow{-4}{*}{\cellcolor{white} RF} & ST-CCT-CAO & 0.896 & \bfseries 0.206 & \bfseries (0.94, 1.00) \\
    \midrule
    \cellcolor{white} & ST-CT & 0.756 & \bfseries 0.052 & \bfseries (0.78, 0.92) \\
    \cellcolor{white} & ST-CCT & 0.795 & \bfseries 0.088 & \bfseries (0.82, 0.95) \\
    \cellcolor{white} & ST-CAO & 0.853 & \bfseries 0.152 & \bfseries (0.94, 1.00) \\
    \multirow{-4}{*}{\cellcolor{white} SVM} & ST-CCT-CAO & 0.866 & \bfseries 0.173 & \bfseries (0.92, 0.99) \\
    \bottomrule
    \end{tabular}
\end{table*}

\begin{table*}[tbp]
    \caption{Model performance on \mbox{DUD-E} datasets using crystal conformer
    queries with input features normalized by mean subtraction and division by
    the standard deviation (``standard'' scaling). We report median AUC, median
    per-dataset $\Delta$AUC, and a sign test 95\% confidence interval (see
    Section~\ref{sec:model_evaluation}). Bold values indicate statistically
    significant confidence intervals that do not include 0.5.}
    \label{table:standard_scaling}
    \centering
    \small
    \rowcolors{2}{lightgray}{}
    \sisetup{detect-weight=true,detect-inline-weight=math}
    \begin{tabular}{ l l S S c }
    \toprule
    Model & Features &
    {\makecell{Median \\ AUC}} &
    {\makecell{Median \\ $\Delta$AUC}} &
    \makecell{Sign Test \\ 95\% CI} \\
    \midrule
    \cellcolor{white} ROCS & TanimotoCombo & 0.697 &  &  \\
    \midrule
    \cellcolor{white} & ST-CT & 0.730 & \bfseries 0.023 & \bfseries (0.72, 0.88) \\
    \cellcolor{white} & ST-CCT & 0.756 & \bfseries 0.059 & \bfseries (0.78, 0.92) \\
    \cellcolor{white} & ST-CAO & 0.814 & \bfseries 0.122 & \bfseries (0.94, 1.00) \\
    \multirow{-4}{*}{\cellcolor{white} LR} & ST-CCT-CAO & 0.824 & \bfseries 0.125 & \bfseries (0.91, 0.99) \\
    \midrule
    \cellcolor{white} & ST-CT & 0.681 & 0.017 & (0.44, 0.64) \\
    \cellcolor{white} & ST-CCT & 0.812 & \bfseries 0.124 & \bfseries (0.91, 0.99) \\
    \cellcolor{white} & ST-CAO & 0.892 & \bfseries 0.201 & \bfseries (0.94, 1.00) \\
    \multirow{-4}{*}{\cellcolor{white} RF} & ST-CCT-CAO & 0.895 & \bfseries 0.213 & \bfseries (0.94, 1.00) \\
    \midrule
    \cellcolor{white} & ST-CT & 0.750 & \bfseries 0.050 & \bfseries (0.78, 0.92) \\
    \cellcolor{white} & ST-CCT & 0.797 & \bfseries 0.095 & \bfseries (0.86, 0.97) \\
    \cellcolor{white} & ST-CAO & 0.867 & \bfseries 0.163 & \bfseries (0.94, 1.00) \\
    \multirow{-4}{*}{\cellcolor{white} SVM} & ST-CCT-CAO & 0.878 & \bfseries 0.178 & \bfseries (0.94, 1.00) \\
    \bottomrule
    \end{tabular}
\end{table*}

\section{Appendix: ROC enrichment}
\label{sec:roc_enrichment}

\subsection{DUD-E}

\begin{table*}[htbp]
    \caption{DUD-E ROC enrichment at 0.005 FPR.}
    \centering
    \small
    \rowcolors{6}{}{lightgray}
    \sisetup{detect-weight=true,detect-inline-weight=math}
    \begin{tabular}{ l l S S c S S c }
    \toprule
     & & \multicolumn{3}{c}{Crystal Conformer} &
         \multicolumn{3}{c}{Generated Conformer} \\
    \cmidrule(lr){3-5} \cmidrule(lr){6-8}
    Model & Features &
    {\makecell{Median \\ $E_{0.005}$}} &
    {\makecell{Median \\ $\Delta E_{0.005}$}} &
    \makecell{Sign Test \\ 95\% CI} &
    {\makecell{Median \\ $E_{0.005}$}} &
    {\makecell{Median \\ $\Delta E_{0.005}$}} &
    \makecell{Sign Test \\ 95\% CI} \\
    \midrule
    \cellcolor{white} ROCS & TanimotoCombo & 28 &  &  & 20 &  &  \\
    \midrule
    \cellcolor{white} & ST-CT & 33 & 0 & (0.43, 0.64) & 25 & 0 & (0.48, 0.70) \\
    \cellcolor{white} & ST-CCT & 23 & -1 & (0.37, 0.58) & 22 & -1 & (0.32, 0.53) \\
    \cellcolor{white} & ST-CAO & 35 & 2 & (0.48, 0.68) & 29 & 2 & (0.49, 0.69) \\
    \multirow{-4}{*}{\cellcolor{white} LR} & ST-CCT-CAO & 30 & 0 & (0.39, 0.60) & 25 & -0 & (0.37, 0.58) \\
    \midrule
    \cellcolor{white} & ST-CT & 33 & \bfseries 3 & \bfseries (0.54, 0.74) & 30 & \bfseries 5 & \bfseries (0.69, 0.87) \\
    \cellcolor{white} & ST-CCT & 75 & \bfseries 36 & \bfseries (0.92, 0.99) & 74 & \bfseries 41 & \bfseries (0.96, 1.00) \\
    \cellcolor{white} & ST-CAO & 116 & \bfseries 79 & \bfseries (0.94, 1.00) & 114 & \bfseries 75 & \bfseries (0.96, 1.00) \\
    \multirow{-4}{*}{\cellcolor{white} RF} & ST-CCT-CAO & 116 & \bfseries 81 & \bfseries (0.94, 1.00) & 118 & \bfseries 80 & \bfseries (0.94, 1.00) \\
    \midrule
    \cellcolor{white} & ST-CT & 28 & \bfseries 1 & \bfseries (0.52, 0.73) & 25 & 0 & (0.45, 0.67) \\
    \cellcolor{white} & ST-CCT & 35 & 1 & (0.48, 0.68) & 29 & \bfseries 2 & \bfseries (0.50, 0.70) \\
    \cellcolor{white} & ST-CAO & 42 & \bfseries 12 & \bfseries (0.67, 0.85) & 41 & \bfseries 17 & \bfseries (0.67, 0.84) \\
    \multirow{-4}{*}{\cellcolor{white} SVM} & ST-CCT-CAO & 50 & \bfseries 15 & \bfseries (0.76, 0.91) & 42 & \bfseries 17 & \bfseries (0.72, 0.88) \\
    \bottomrule
    \end{tabular}
\end{table*}

\begin{table*}[htbp]
    \caption{DUD-E ROC enrichment at 0.01 FPR.}
    \centering
    \small
    \rowcolors{6}{}{lightgray}
    \sisetup{detect-weight=true,detect-inline-weight=math}
    \begin{tabular}{ l l S S c S S c }
    \toprule
     & & \multicolumn{3}{c}{Crystal Conformer} &
         \multicolumn{3}{c}{Generated Conformer} \\
    \cmidrule(lr){3-5} \cmidrule(lr){6-8}
    Model & Features &
    {\makecell{Median \\ $E_{0.01}$}} &
    {\makecell{Median \\ $\Delta E_{0.01}$}} &
    \makecell{Sign Test \\ 95\% CI} &
    {\makecell{Median \\ $E_{0.01}$}} &
    {\makecell{Median \\ $\Delta E_{0.01}$}} &
    \makecell{Sign Test \\ 95\% CI} \\
    \midrule
    \cellcolor{white} ROCS & TanimotoCombo & 16 &  &  & 12 &  &  \\
    \midrule
    \cellcolor{white} & ST-CT & 20 & 0 & (0.44, 0.66) & 16 & \bfseries 1 & \bfseries (0.52, 0.74) \\
    \cellcolor{white} & ST-CCT & 18 & 0 & (0.43, 0.63) & 17 & 0 & (0.41, 0.61) \\
    \cellcolor{white} & ST-CAO & 23 & \bfseries 3 & \bfseries (0.55, 0.75) & 22 & \bfseries 4 & \bfseries (0.61, 0.80) \\
    \multirow{-4}{*}{\cellcolor{white} LR} & ST-CCT-CAO & 24 & \bfseries 3 & \bfseries (0.53, 0.73) & 22 & \bfseries 3 & \bfseries (0.56, 0.76) \\
    \midrule
    \cellcolor{white} & ST-CT & 20 & \bfseries 2 & \bfseries (0.53, 0.73) & 19 & \bfseries 3 & \bfseries (0.64, 0.82) \\
    \cellcolor{white} & ST-CCT & 42 & \bfseries 21 & \bfseries (0.94, 1.00) & 41 & \bfseries 23 & \bfseries (0.92, 0.99) \\
    \cellcolor{white} & ST-CAO & 62 & \bfseries 41 & \bfseries (0.94, 1.00) & 62 & \bfseries 40 & \bfseries (0.94, 1.00) \\
    \multirow{-4}{*}{\cellcolor{white} RF} & ST-CCT-CAO & 64 & \bfseries 42 & \bfseries (0.94, 1.00) & 62 & \bfseries 42 & \bfseries (0.94, 1.00) \\
    \midrule
    \cellcolor{white} & ST-CT & 19 & \bfseries 1 & \bfseries (0.51, 0.72) & 16 & \bfseries 1 & \bfseries (0.56, 0.76) \\
    \cellcolor{white} & ST-CCT & 24 & \bfseries 2 & \bfseries (0.58, 0.77) & 20 & \bfseries 3 & \bfseries (0.56, 0.75) \\
    \cellcolor{white} & ST-CAO & 29 & \bfseries 8 & \bfseries (0.79, 0.93) & 28 & \bfseries 11 & \bfseries (0.76, 0.91) \\
    \multirow{-4}{*}{\cellcolor{white} SVM} & ST-CCT-CAO & 33 & \bfseries 11 & \bfseries (0.78, 0.92) & 30 & \bfseries 13 & \bfseries (0.83, 0.95) \\
    \bottomrule
    \end{tabular}
\end{table*}

\begin{table*}[htbp]
    \caption{DUD-E ROC enrichment at 0.02 FPR.}
    \centering
    \small
    \rowcolors{6}{}{lightgray}
    \sisetup{detect-weight=true,detect-inline-weight=math}
    \begin{tabular}{ l l S S c S S c }
    \toprule
     & & \multicolumn{3}{c}{Crystal Conformer} &
         \multicolumn{3}{c}{Generated Conformer} \\
    \cmidrule(lr){3-5} \cmidrule(lr){6-8}
    Model & Features &
    {\makecell{Median \\ $E_{0.02}$}} &
    {\makecell{Median \\ $\Delta E_{0.02}$}} &
    \makecell{Sign Test \\ 95\% CI} &
    {\makecell{Median \\ $E_{0.02}$}} &
    {\makecell{Median \\ $\Delta E_{0.02}$}} &
    \makecell{Sign Test \\ 95\% CI} \\
    \midrule
    \cellcolor{white} ROCS & TanimotoCombo & 10 &  &  & 9 &  &  \\
    \midrule
    \cellcolor{white} & ST-CT & 12 & \bfseries 0 & \bfseries (0.51, 0.72) & 10 & 0 & (0.49, 0.70) \\
    \cellcolor{white} & ST-CCT & 13 & 1 & (0.49, 0.69) & 12 & \bfseries 1 & \bfseries (0.51, 0.71) \\
    \cellcolor{white} & ST-CAO & 15 & \bfseries 3 & \bfseries (0.68, 0.85) & 14 & \bfseries 3 & \bfseries (0.72, 0.88) \\
    \multirow{-4}{*}{\cellcolor{white} LR} & ST-CCT-CAO & 16 & \bfseries 3 & \bfseries (0.71, 0.87) & 14 & \bfseries 3 & \bfseries (0.72, 0.88) \\
    \midrule
    \cellcolor{white} & ST-CT & 12 & \bfseries 1 & \bfseries (0.50, 0.71) & 12 & \bfseries 2 & \bfseries (0.64, 0.82) \\
    \cellcolor{white} & ST-CCT & 24 & \bfseries 11 & \bfseries (0.94, 1.00) & 23 & \bfseries 12 & \bfseries (0.91, 0.99) \\
    \cellcolor{white} & ST-CAO & 34 & \bfseries 21 & \bfseries (0.94, 1.00) & 33 & \bfseries 21 & \bfseries (0.96, 1.00) \\
    \multirow{-4}{*}{\cellcolor{white} RF} & ST-CCT-CAO & 35 & \bfseries 21 & \bfseries (0.96, 1.00) & 33 & \bfseries 21 & \bfseries (0.94, 1.00) \\
    \midrule
    \cellcolor{white} & ST-CT & 13 & \bfseries 1 & \bfseries (0.54, 0.74) & 12 & \bfseries 1 & \bfseries (0.54, 0.74) \\
    \cellcolor{white} & ST-CCT & 16 & \bfseries 2 & \bfseries (0.72, 0.88) & 13 & \bfseries 2 & \bfseries (0.71, 0.88) \\
    \cellcolor{white} & ST-CAO & 19 & \bfseries 8 & \bfseries (0.82, 0.95) & 20 & \bfseries 8 & \bfseries (0.86, 0.97) \\
    \multirow{-4}{*}{\cellcolor{white} SVM} & ST-CCT-CAO & 21 & \bfseries 8 & \bfseries (0.86, 0.97) & 20 & \bfseries 10 & \bfseries (0.88, 0.98) \\
    \bottomrule
    \end{tabular}
\end{table*}

\begin{table*}[htbp]
    \caption{DUD-E ROC enrichment at 0.05 FPR.}
    \centering
    \small
    \rowcolors{6}{}{lightgray}
    \sisetup{detect-weight=true,detect-inline-weight=math}
    \begin{tabular}{ l l S S c S S c }
    \toprule
     & & \multicolumn{3}{c}{Crystal Conformer} &
         \multicolumn{3}{c}{Generated Conformer} \\
    \cmidrule(lr){3-5} \cmidrule(lr){6-8}
    Model & Features &
    {\makecell{Median \\ $E_{0.05}$}} &
    {\makecell{Median \\ $\Delta E_{0.05}$}} &
    \makecell{Sign Test \\ 95\% CI} &
    {\makecell{Median \\ $E_{0.05}$}} &
    {\makecell{Median \\ $\Delta E_{0.05}$}} &
    \makecell{Sign Test \\ 95\% CI} \\
    \midrule
    \cellcolor{white} ROCS & TanimotoCombo & 6 &  &  & 4 &  &  \\
    \midrule
    \cellcolor{white} & ST-CT & 7 & \bfseries 1 & \bfseries (0.63, 0.82) & 6 & \bfseries 0 & \bfseries (0.62, 0.82) \\
    \cellcolor{white} & ST-CCT & 7 & \bfseries 1 & \bfseries (0.59, 0.78) & 7 & \bfseries 1 & \bfseries (0.62, 0.81) \\
    \cellcolor{white} & ST-CAO & 9 & \bfseries 2 & \bfseries (0.83, 0.95) & 8 & \bfseries 2 & \bfseries (0.87, 0.98) \\
    \multirow{-4}{*}{\cellcolor{white} LR} & ST-CCT-CAO & 9 & \bfseries 2 & \bfseries (0.82, 0.95) & 9 & \bfseries 2 & \bfseries (0.89, 0.98) \\
    \midrule
    \cellcolor{white} & ST-CT & 7 & \bfseries 1 & \bfseries (0.51, 0.71) & 6 & \bfseries 1 & \bfseries (0.63, 0.81) \\
    \cellcolor{white} & ST-CCT & 11 & \bfseries 5 & \bfseries (0.92, 0.99) & 11 & \bfseries 5 & \bfseries (0.91, 0.99) \\
    \cellcolor{white} & ST-CAO & 15 & \bfseries 8 & \bfseries (0.96, 1.00) & 14 & \bfseries 9 & \bfseries (0.94, 1.00) \\
    \multirow{-4}{*}{\cellcolor{white} RF} & ST-CCT-CAO & 15 & \bfseries 8 & \bfseries (0.96, 1.00) & 14 & \bfseries 9 & \bfseries (0.96, 1.00) \\
    \midrule
    \cellcolor{white} & ST-CT & 7 & \bfseries 1 & \bfseries (0.67, 0.85) & 7 & \bfseries 1 & \bfseries (0.71, 0.88) \\
    \cellcolor{white} & ST-CCT & 8 & \bfseries 2 & \bfseries (0.75, 0.90) & 8 & \bfseries 2 & \bfseries (0.78, 0.92) \\
    \cellcolor{white} & ST-CAO & 11 & \bfseries 5 & \bfseries (0.85, 0.96) & 11 & \bfseries 5 & \bfseries (0.92, 0.99) \\
    \multirow{-4}{*}{\cellcolor{white} SVM} & ST-CCT-CAO & 12 & \bfseries 6 & \bfseries (0.88, 0.98) & 12 & \bfseries 6 & \bfseries (0.88, 0.98) \\
    \bottomrule
    \end{tabular}
\end{table*}

\subsection{MUV}

\begin{table*}[htbp]
    \caption{MUV ROC enrichment at 0.005 FPR.}
    \centering
    \small
    \rowcolors{2}{lightgray}{}
    \sisetup{detect-weight=true,detect-inline-weight=math}
    \begin{tabular}{ l l S S c }
    \toprule
    Model & Features &
    {\makecell{Median \\ $E_{0.005}$}} &
    {\makecell{Median \\ $\Delta E_{0.005}$}} &
    \makecell{Sign Test \\ 95\% CI} \\
    \midrule
    \cellcolor{white} ROCS & TanimotoCombo & 0 &  &  \\
    \midrule
    \cellcolor{white} & ST-CT & 0 & 0 & (0.41, 0.60) \\
    \cellcolor{white} & ST-CCT & 0 & \bfseries \color{red} 0 & \bfseries \color{red} (0.30, 0.45) \\
    \cellcolor{white} & ST-CAO & 0 & 0 & (0.38, 0.53) \\
    \multirow{-4}{*}{\cellcolor{white} LR} & ST-CCT-CAO & 0 & 0 & (0.36, 0.50) \\
    \midrule
    \cellcolor{white} & ST-CT & 0 & \bfseries \color{red} 0 & \bfseries \color{red} (0.34, 0.48) \\
    \cellcolor{white} & ST-CCT & 1 & \bfseries 0 & \bfseries (0.60, 0.72) \\
    \cellcolor{white} & ST-CAO & 10 & \bfseries 7 & \bfseries (0.70, 0.80) \\
    \multirow{-4}{*}{\cellcolor{white} RF} & ST-CCT-CAO & 12 & \bfseries 8 & \bfseries (0.70, 0.80) \\
    \midrule
    \cellcolor{white} & ST-CT & 0 & \bfseries \color{red} 0 & \bfseries \color{red} (0.33, 0.49) \\
    \cellcolor{white} & ST-CCT & 0 & \bfseries \color{red} 0 & \bfseries \color{red} (0.26, 0.40) \\
    \cellcolor{white} & ST-CAO & 0 & 0 & (0.38, 0.51) \\
    \multirow{-4}{*}{\cellcolor{white} SVM} & ST-CCT-CAO & 0 & 0 & (0.39, 0.53) \\
    \bottomrule
    \end{tabular}
\end{table*}

\begin{table*}[htbp]
    \caption{MUV ROC enrichment at 0.01 FPR.}
    \centering
    \small
    \rowcolors{2}{lightgray}{}
    \sisetup{detect-weight=true,detect-inline-weight=math}
    \begin{tabular}{ l l S S c }
    \toprule
    Model & Features &
    {\makecell{Median \\ $E_{0.01}$}} &
    {\makecell{Median \\ $\Delta E_{0.01}$}} &
    \makecell{Sign Test \\ 95\% CI} \\
    \midrule
    \cellcolor{white} ROCS & TanimotoCombo & 4 &  &  \\
    \midrule
    \cellcolor{white} & ST-CT & 0 & 0 & (0.34, 0.51) \\
    \cellcolor{white} & ST-CCT & 0 & \bfseries \color{red} 0 & \bfseries \color{red} (0.34, 0.47) \\
    \cellcolor{white} & ST-CAO & 4 & 0 & (0.43, 0.57) \\
    \multirow{-4}{*}{\cellcolor{white} LR} & ST-CCT-CAO & 4 & 0 & (0.44, 0.57) \\
    \midrule
    \cellcolor{white} & ST-CT & 0 & \bfseries \color{red} 0 & \bfseries \color{red} (0.35, 0.48) \\
    \cellcolor{white} & ST-CCT & 4 & \bfseries 0 & \bfseries (0.55, 0.67) \\
    \cellcolor{white} & ST-CAO & 7 & \bfseries 5 & \bfseries (0.67, 0.77) \\
    \multirow{-4}{*}{\cellcolor{white} RF} & ST-CCT-CAO & 8 & \bfseries 5 & \bfseries (0.70, 0.80) \\
    \midrule
    \cellcolor{white} & ST-CT & 0 & \bfseries \color{red} 0 & \bfseries \color{red} (0.29, 0.43) \\
    \cellcolor{white} & ST-CCT & 0 & \bfseries \color{red} 0 & \bfseries \color{red} (0.29, 0.41) \\
    \cellcolor{white} & ST-CAO & 4 & 0 & (0.45, 0.57) \\
    \multirow{-4}{*}{\cellcolor{white} SVM} & ST-CCT-CAO & 4 & 0 & (0.45, 0.57) \\
    \bottomrule
    \end{tabular}
\end{table*}

\begin{table*}[htbp]
    \caption{MUV ROC enrichment at 0.02 FPR.}
    \centering
    \small
    \rowcolors{2}{lightgray}{}
    \sisetup{detect-weight=true,detect-inline-weight=math}
    \begin{tabular}{ l l S S c }
    \toprule
    Model & Features &
    {\makecell{Median \\ $E_{0.02}$}} &
    {\makecell{Median \\ $\Delta E_{0.02}$}} &
    \makecell{Sign Test \\ 95\% CI} \\
    \midrule
    \cellcolor{white} ROCS & TanimotoCombo & 2 &  &  \\
    \midrule
    \cellcolor{white} & ST-CT & 2 & 0 & (0.38, 0.52) \\
    \cellcolor{white} & ST-CCT & 2 & 0 & (0.44, 0.57) \\
    \cellcolor{white} & ST-CAO & 2 & \bfseries 0 & \bfseries (0.51, 0.63) \\
    \multirow{-4}{*}{\cellcolor{white} LR} & ST-CCT-CAO & 2 & 0 & (0.49, 0.61) \\
    \midrule
    \cellcolor{white} & ST-CT & 2 & 0 & (0.40, 0.51) \\
    \cellcolor{white} & ST-CCT & 3 & \bfseries 0 & \bfseries (0.58, 0.69) \\
    \cellcolor{white} & ST-CAO & 4 & \bfseries 2 & \bfseries (0.66, 0.76) \\
    \multirow{-4}{*}{\cellcolor{white} RF} & ST-CCT-CAO & 5 & \bfseries 3 & \bfseries (0.69, 0.78) \\
    \midrule
    \cellcolor{white} & ST-CT & 2 & 0 & (0.40, 0.52) \\
    \cellcolor{white} & ST-CCT & 2 & 0 & (0.38, 0.50) \\
    \cellcolor{white} & ST-CAO & 2 & \bfseries 0 & \bfseries (0.54, 0.65) \\
    \multirow{-4}{*}{\cellcolor{white} SVM} & ST-CCT-CAO & 2 & \bfseries 0 & \bfseries (0.54, 0.66) \\
    \bottomrule
    \end{tabular}
\end{table*}

\begin{table*}[htbp]
    \caption{MUV ROC enrichment at 0.05 FPR.}
    \centering
    \small
    \rowcolors{2}{lightgray}{}
    \sisetup{detect-weight=true,detect-inline-weight=math}
    \begin{tabular}{ l l S S c }
    \toprule
    Model & Features &
    {\makecell{Median \\ $E_{0.05}$}} &
    {\makecell{Median \\ $\Delta E_{0.05}$}} &
    \makecell{Sign Test \\ 95\% CI} \\
    \midrule
    \cellcolor{white} ROCS & TanimotoCombo & 2 &  &  \\
    \midrule
    \cellcolor{white} & ST-CT & 2 & 0 & (0.42, 0.55) \\
    \cellcolor{white} & ST-CCT & 2 & 0 & (0.42, 0.53) \\
    \cellcolor{white} & ST-CAO & 2 & \bfseries 0 & \bfseries (0.53, 0.63) \\
    \multirow{-4}{*}{\cellcolor{white} LR} & ST-CCT-CAO & 2 & \bfseries 0 & \bfseries (0.55, 0.66) \\
    \midrule
    \cellcolor{white} & ST-CT & 2 & -1 & (0.41, 0.51) \\
    \cellcolor{white} & ST-CCT & 2 & \bfseries 0 & \bfseries (0.56, 0.66) \\
    \cellcolor{white} & ST-CAO & 3 & \bfseries 1 & \bfseries (0.64, 0.73) \\
    \multirow{-4}{*}{\cellcolor{white} RF} & ST-CCT-CAO & 3 & \bfseries 1 & \bfseries (0.63, 0.73) \\
    \midrule
    \cellcolor{white} & ST-CT & 2 & 0 & (0.40, 0.51) \\
    \cellcolor{white} & ST-CCT & 2 & 0 & (0.42, 0.53) \\
    \cellcolor{white} & ST-CAO & 3 & \bfseries 1 & \bfseries (0.58, 0.69) \\
    \multirow{-4}{*}{\cellcolor{white} SVM} & ST-CCT-CAO & 3 & \bfseries 1 & \bfseries (0.60, 0.70) \\
    \bottomrule
    \end{tabular}
\end{table*}

\subsection{ChEMBL}

\begin{table*}[htbp]
    \caption{ChEMBL ROC enrichment at 0.005 FPR.}
    \centering
    \small
    \rowcolors{2}{lightgray}{}
    \sisetup{detect-weight=true,detect-inline-weight=math}
    \begin{tabular}{ l l S S c }
    \toprule
    Model & Features &
    {\makecell{Median \\ $E_{0.005}$}} &
    {\makecell{Median \\ $\Delta E_{0.005}$}} &
    \makecell{Sign Test \\ 95\% CI} \\
    \midrule
    \cellcolor{white} ROCS & TanimotoCombo & 19 &  &  \\
    \midrule
    \cellcolor{white} & ST-CT & 19 & 0 & (0.49, 0.52) \\
    \cellcolor{white} & ST-CCT & 16 & \bfseries \color{red} -0 & \bfseries \color{red} (0.40, 0.43) \\
    \cellcolor{white} & ST-CAO & 21 & \bfseries 3 & \bfseries (0.59, 0.62) \\
    \multirow{-4}{*}{\cellcolor{white} LR} & ST-CCT-CAO & 21 & \bfseries 3 & \bfseries (0.56, 0.59) \\
    \midrule
    \cellcolor{white} & ST-CT & 17 & \bfseries \color{red} 0 & \bfseries \color{red} (0.43, 0.46) \\
    \cellcolor{white} & ST-CCT & 34 & \bfseries 13 & \bfseries (0.81, 0.84) \\
    \cellcolor{white} & ST-CAO & 60 & \bfseries 36 & \bfseries (0.92, 0.94) \\
    \multirow{-4}{*}{\cellcolor{white} RF} & ST-CCT-CAO & 62 & \bfseries 38 & \bfseries (0.93, 0.94) \\
    \midrule
    \cellcolor{white} & ST-CT & 20 & \bfseries 0 & \bfseries (0.51, 0.55) \\
    \cellcolor{white} & ST-CCT & 19 & 0 & (0.49, 0.53) \\
    \cellcolor{white} & ST-CAO & 21 & \bfseries 3 & \bfseries (0.57, 0.60) \\
    \multirow{-4}{*}{\cellcolor{white} SVM} & ST-CCT-CAO & 27 & \bfseries 7 & \bfseries (0.69, 0.72) \\
    \bottomrule
    \end{tabular}
\end{table*}

\begin{table*}[htbp]
    \caption{ChEMBL ROC enrichment at 0.01 FPR.}
    \centering
    \small
    \rowcolors{2}{lightgray}{}
    \sisetup{detect-weight=true,detect-inline-weight=math}
    \begin{tabular}{ l l S S c }
    \toprule
    Model & Features &
    {\makecell{Median \\ $E_{0.01}$}} &
    {\makecell{Median \\ $\Delta E_{0.01}$}} &
    \makecell{Sign Test \\ 95\% CI} \\
    \midrule
    \cellcolor{white} ROCS & TanimotoCombo & 11 &  &  \\
    \midrule
    \cellcolor{white} & ST-CT & 12 & \bfseries 0 & \bfseries (0.52, 0.55) \\
    \cellcolor{white} & ST-CCT & 12 & \bfseries \color{red} 0 & \bfseries \color{red} (0.46, 0.49) \\
    \cellcolor{white} & ST-CAO & 16 & \bfseries 4 & \bfseries (0.68, 0.71) \\
    \multirow{-4}{*}{\cellcolor{white} LR} & ST-CCT-CAO & 18 & \bfseries 5 & \bfseries (0.69, 0.72) \\
    \midrule
    \cellcolor{white} & ST-CT & 10 & \bfseries \color{red} 0 & \bfseries \color{red} (0.45, 0.49) \\
    \cellcolor{white} & ST-CCT & 22 & \bfseries 9 & \bfseries (0.83, 0.86) \\
    \cellcolor{white} & ST-CAO & 36 & \bfseries 21 & \bfseries (0.94, 0.95) \\
    \multirow{-4}{*}{\cellcolor{white} RF} & ST-CCT-CAO & 38 & \bfseries 23 & \bfseries (0.95, 0.96) \\
    \midrule
    \cellcolor{white} & ST-CT & 13 & \bfseries 0 & \bfseries (0.57, 0.60) \\
    \cellcolor{white} & ST-CCT & 13 & \bfseries 1 & \bfseries (0.57, 0.61) \\
    \cellcolor{white} & ST-CAO & 17 & \bfseries 4 & \bfseries (0.70, 0.73) \\
    \multirow{-4}{*}{\cellcolor{white} SVM} & ST-CCT-CAO & 20 & \bfseries 8 & \bfseries (0.80, 0.82) \\
    \bottomrule
    \end{tabular}
\end{table*}

\begin{table*}[htbp]
    \caption{ChEMBL ROC enrichment at 0.02 FPR.}
    \centering
    \small
    \rowcolors{2}{lightgray}{}
    \sisetup{detect-weight=true,detect-inline-weight=math}
    \begin{tabular}{ l l S S c }
    \toprule
    Model & Features &
    {\makecell{Median \\ $E_{0.02}$}} &
    {\makecell{Median \\ $\Delta E_{0.02}$}} &
    \makecell{Sign Test \\ 95\% CI} \\
    \midrule
    \cellcolor{white} ROCS & TanimotoCombo & 7 &  &  \\
    \midrule
    \cellcolor{white} & ST-CT & 8 & \bfseries 0 & \bfseries (0.57, 0.60) \\
    \cellcolor{white} & ST-CCT & 8 & \bfseries 0 & \bfseries (0.52, 0.55) \\
    \cellcolor{white} & ST-CAO & 11 & \bfseries 3 & \bfseries (0.77, 0.80) \\
    \multirow{-4}{*}{\cellcolor{white} LR} & ST-CCT-CAO & 13 & \bfseries 5 & \bfseries (0.82, 0.85) \\
    \midrule
    \cellcolor{white} & ST-CT & 7 & 0 & (0.48, 0.51) \\
    \cellcolor{white} & ST-CCT & 14 & \bfseries 6 & \bfseries (0.85, 0.87) \\
    \cellcolor{white} & ST-CAO & 22 & \bfseries 12 & \bfseries (0.95, 0.96) \\
    \multirow{-4}{*}{\cellcolor{white} RF} & ST-CCT-CAO & 22 & \bfseries 13 & \bfseries (0.96, 0.97) \\
    \midrule
    \cellcolor{white} & ST-CT & 9 & \bfseries 1 & \bfseries (0.64, 0.67) \\
    \cellcolor{white} & ST-CCT & 10 & \bfseries 2 & \bfseries (0.66, 0.69) \\
    \cellcolor{white} & ST-CAO & 13 & \bfseries 5 & \bfseries (0.82, 0.85) \\
    \multirow{-4}{*}{\cellcolor{white} SVM} & ST-CCT-CAO & 15 & \bfseries 7 & \bfseries (0.88, 0.90) \\
    \bottomrule
    \end{tabular}
\end{table*}

\begin{table*}[htbp]
    \caption{ChEMBL ROC enrichment at 0.05 FPR.}
    \centering
    \small
    \rowcolors{2}{lightgray}{}
    \sisetup{detect-weight=true,detect-inline-weight=math}
    \begin{tabular}{ l l S S c }
    \toprule
    Model & Features &
    {\makecell{Median \\ $E_{0.05}$}} &
    {\makecell{Median \\ $\Delta E_{0.05}$}} &
    \makecell{Sign Test \\ 95\% CI} \\
    \midrule
    \cellcolor{white} ROCS & TanimotoCombo & 4 &  &  \\
    \midrule
    \cellcolor{white} & ST-CT & 5 & \bfseries 0 & \bfseries (0.61, 0.64) \\
    \cellcolor{white} & ST-CCT & 5 & \bfseries 1 & \bfseries (0.64, 0.67) \\
    \cellcolor{white} & ST-CAO & 7 & \bfseries 3 & \bfseries (0.87, 0.89) \\
    \multirow{-4}{*}{\cellcolor{white} LR} & ST-CCT-CAO & 8 & \bfseries 4 & \bfseries (0.91, 0.93) \\
    \midrule
    \cellcolor{white} & ST-CT & 4 & \bfseries 0 & \bfseries (0.53, 0.56) \\
    \cellcolor{white} & ST-CCT & 8 & \bfseries 3 & \bfseries (0.86, 0.88) \\
    \cellcolor{white} & ST-CAO & 11 & \bfseries 6 & \bfseries (0.96, 0.97) \\
    \multirow{-4}{*}{\cellcolor{white} RF} & ST-CCT-CAO & 11 & \bfseries 6 & \bfseries (0.97, 0.98) \\
    \midrule
    \cellcolor{white} & ST-CT & 5 & \bfseries 1 & \bfseries (0.72, 0.75) \\
    \cellcolor{white} & ST-CCT & 6 & \bfseries 2 & \bfseries (0.78, 0.80) \\
    \cellcolor{white} & ST-CAO & 8 & \bfseries 4 & \bfseries (0.90, 0.92) \\
    \multirow{-4}{*}{\cellcolor{white} SVM} & ST-CCT-CAO & 9 & \bfseries 5 & \bfseries (0.92, 0.94) \\
    \bottomrule
    \end{tabular}
\end{table*}


\section{Appendix: Tversky features}
\label{sec:tversky}

The tables in this section report ROC AUC and enrichment scores for models built
using reference Tversky scores for shape, color, and color components. This is
in contrast to the default approach of using Tanimoto similarity. Feature
abbreviations are given in \tablename~\ref{table:abbreviations}. Note that color
atom overlap features are the same for Tanimoto and Tversky models.

\subsection{ROC AUC}

\begin{table*}[htbp]
    \caption{DUD-E results.}
    \label{table:dude_tversky}
    \centering
    \small
    \rowcolors{6}{}{lightgray}
    \sisetup{detect-weight=true,detect-inline-weight=math}
    \begin{tabular}{ l l S S c S S c }
    \toprule
     & & \multicolumn{3}{c}{Crystal Conformer} &
         \multicolumn{3}{c}{Generated Conformer} \\
    \cmidrule(lr){3-5} \cmidrule(lr){6-8}
    Model & Features &
    {\makecell{Median \\ AUC}} &
    {\makecell{Median \\ $\Delta$AUC}} &
    \makecell{Sign Test \\ 95\% CI} &
    {\makecell{Median \\ AUC}} &
    {\makecell{Median \\ $\Delta$AUC}} &
    \makecell{Sign Test \\ 95\% CI} \\
    \midrule
    \cellcolor{white} ROCS & TverskyCombo & 0.749 &  &  & 0.724 &  &  \\
    \midrule
    \cellcolor{white} & STv-CTv & 0.756 & 0.003 & (0.50, 0.70) & 0.734 & \bfseries 0.009 & \bfseries (0.66, 0.83) \\
    \cellcolor{white} & STv-CCTv & 0.787 & \bfseries 0.033 & \bfseries (0.77, 0.91) & 0.772 & \bfseries 0.039 & \bfseries (0.76, 0.91) \\
    \cellcolor{white} & ST-CAO & 0.813 & \bfseries 0.058 & \bfseries (0.77, 0.91) & 0.813 & \bfseries 0.073 & \bfseries (0.81, 0.94) \\
    \multirow{-4}{*}{\cellcolor{white} LR} & STv-CCTv-CAO & 0.832 & \bfseries 0.073 & \bfseries (0.86, 0.97) & 0.824 & \bfseries 0.086 & \bfseries (0.86, 0.97) \\
    \midrule
    \cellcolor{white} & STv-CTv & 0.727 & \bfseries \color{red} -0.017 & \bfseries \color{red} (0.26, 0.46) & 0.709 & -0.006 & (0.36, 0.56) \\
    \cellcolor{white} & STv-CCTv & 0.829 & \bfseries 0.063 & \bfseries (0.82, 0.95) & 0.816 & \bfseries 0.080 & \bfseries (0.82, 0.95) \\
    \cellcolor{white} & ST-CAO & 0.893 & \bfseries 0.134 & \bfseries (0.89, 0.98) & 0.888 & \bfseries 0.147 & \bfseries (0.91, 0.99) \\
    \multirow{-4}{*}{\cellcolor{white} RF} & STv-CCTv-CAO & 0.896 & \bfseries 0.134 & \bfseries (0.91, 0.99) & 0.891 & \bfseries 0.149 & \bfseries (0.91, 0.99) \\
    \midrule
    \cellcolor{white} & STv-CTv & 0.770 & \bfseries 0.013 & \bfseries (0.58, 0.77) & 0.745 & \bfseries 0.021 & \bfseries (0.72, 0.88) \\
    \cellcolor{white} & STv-CCTv & 0.806 & \bfseries 0.052 & \bfseries (0.82, 0.95) & 0.803 & \bfseries 0.062 & \bfseries (0.85, 0.96) \\
    \cellcolor{white} & ST-CAO & 0.864 & \bfseries 0.101 & \bfseries (0.88, 0.98) & 0.856 & \bfseries 0.120 & \bfseries (0.88, 0.98) \\
    \multirow{-4}{*}{\cellcolor{white} SVM} & STv-CCTv-CAO & 0.858 & \bfseries 0.100 & \bfseries (0.89, 0.98) & 0.864 & \bfseries 0.123 & \bfseries (0.91, 0.99) \\
    \bottomrule
    \end{tabular}
\end{table*}

\begin{table*}[htbp]
    \caption{MUV results.}
    \label{table:muv_tversky}
    \centering
    \small
    \rowcolors{2}{lightgray}{}
    \sisetup{detect-weight=true,detect-inline-weight=math}
    \begin{tabular}{ l l S S c }
    \toprule
    Model & Features &
    {\makecell{Median \\ AUC}} &
    {\makecell{Median \\ $\Delta$AUC}} &
    \makecell{Sign Test \\ 95\% CI} \\
    \midrule
    \cellcolor{white} ROCS & TverskyCombo & 0.601 &  &  \\
    \midrule
    \cellcolor{white} & STv-CTv & 0.618 & -0.002 & (0.42, 0.52) \\
    \cellcolor{white} & STv-CCTv & 0.608 & -0.005 & (0.42, 0.52) \\
    \cellcolor{white} & ST-CAO & 0.615 & \bfseries 0.013 & \bfseries (0.52, 0.62) \\
    \multirow{-4}{*}{\cellcolor{white} LR} & STv-CCTv-CAO & 0.621 & \bfseries 0.017 & \bfseries (0.51, 0.61) \\
    \midrule
    \cellcolor{white} & STv-CTv & 0.522 & \bfseries \color{red} -0.074 & \bfseries \color{red} (0.19, 0.27) \\
    \cellcolor{white} & STv-CCTv & 0.539 & \bfseries \color{red} -0.050 & \bfseries \color{red} (0.27, 0.36) \\
    \cellcolor{white} & ST-CAO & 0.562 & \bfseries \color{red} -0.020 & \bfseries \color{red} (0.37, 0.47) \\
    \multirow{-4}{*}{\cellcolor{white} RF} & STv-CCTv-CAO & 0.563 & \bfseries \color{red} -0.031 & \bfseries \color{red} (0.36, 0.45) \\
    \midrule
    \cellcolor{white} & STv-CTv & 0.608 & \bfseries \color{red} -0.010 & \bfseries \color{red} (0.40, 0.49) \\
    \cellcolor{white} & STv-CCTv & 0.591 & \bfseries \color{red} -0.012 & \bfseries \color{red} (0.40, 0.50) \\
    \cellcolor{white} & ST-CAO & 0.609 & 0.018 & (0.49, 0.59) \\
    \multirow{-4}{*}{\cellcolor{white} SVM} & STv-CCTv-CAO & 0.607 & 0.012 & (0.48, 0.58) \\
    \bottomrule
    \end{tabular}
\end{table*}

\begin{table*}[htbp]
    \caption{ChEMBL results.}
    \label{table:chembl_tversky}
    \centering
    \small
    \rowcolors{2}{lightgray}{}
    \sisetup{detect-weight=true,detect-inline-weight=math}
    \begin{tabular}{ l l S S c }
    \toprule
    Model & Features &
    {\makecell{Median \\ AUC}} &
    {\makecell{Median \\ $\Delta$AUC}} &
    \makecell{Sign Test \\ 95\% CI} \\
    \midrule
    \cellcolor{white} ROCS & TverskyCombo & 0.688 &  &  \\
    \midrule
    \cellcolor{white} & STv-CTv & 0.728 & \bfseries 0.010 & \bfseries (0.65, 0.68) \\
    \cellcolor{white} & STv-CCTv & 0.771 & \bfseries 0.049 & \bfseries (0.81, 0.84) \\
    \cellcolor{white} & ST-CAO & 0.783 & \bfseries 0.076 & \bfseries (0.75, 0.78) \\
    \multirow{-4}{*}{\cellcolor{white} LR} & STv-CCTv-CAO & 0.834 & \bfseries 0.121 & \bfseries (0.93, 0.94) \\
    \midrule
    \cellcolor{white} & STv-CTv & 0.659 & \bfseries \color{red} -0.033 & \bfseries \color{red} (0.36, 0.39) \\
    \cellcolor{white} & STv-CCTv & 0.763 & \bfseries 0.051 & \bfseries (0.67, 0.70) \\
    \cellcolor{white} & ST-CAO & 0.821 & \bfseries 0.111 & \bfseries (0.84, 0.86) \\
    \multirow{-4}{*}{\cellcolor{white} RF} & STv-CCTv-CAO & 0.826 & \bfseries 0.113 & \bfseries (0.88, 0.90) \\
    \midrule
    \cellcolor{white} & STv-CTv & 0.749 & \bfseries 0.026 & \bfseries (0.73, 0.76) \\
    \cellcolor{white} & STv-CCTv & 0.781 & \bfseries 0.062 & \bfseries (0.81, 0.84) \\
    \cellcolor{white} & ST-CAO & 0.804 & \bfseries 0.095 & \bfseries (0.80, 0.83) \\
    \multirow{-4}{*}{\cellcolor{white} SVM} & STv-CCTv-CAO & 0.814 & \bfseries 0.102 & \bfseries (0.87, 0.89) \\
    \bottomrule
    \end{tabular}
\end{table*}

\subsection{ROC enrichment}

\subsubsection{DUD-E}

\begin{table*}[htbp]
    \caption{DUD-E ROC enrichment at 0.005 FPR.}
    \centering
    \small
    \rowcolors{6}{}{lightgray}
    \sisetup{detect-weight=true,detect-inline-weight=math}
    \begin{tabular}{ l l S S c S S c }
    \toprule
     & & \multicolumn{3}{c}{Crystal Conformer} &
         \multicolumn{3}{c}{Generated Conformer} \\
    \cmidrule(lr){3-5} \cmidrule(lr){6-8}
    Model & Features &
    {\makecell{Median \\ $E_{0.005}$}} &
    {\makecell{Median \\ $\Delta E_{0.005}$}} &
    \makecell{Sign Test \\ 95\% CI} &
    {\makecell{Median \\ $E_{0.005}$}} &
    {\makecell{Median \\ $\Delta E_{0.005}$}} &
    \makecell{Sign Test \\ 95\% CI} \\
    \midrule
    \cellcolor{white} ROCS & TverskyCombo & 34 &  &  & 25 &  &  \\
    \midrule
    \cellcolor{white} & STv-CTv & 32 & 0 & (0.30, 0.54) & 24 & 0 & (0.31, 0.54) \\
    \cellcolor{white} & STv-CCTv & 39 & 0 & (0.38, 0.59) & 30 & 0 & (0.40, 0.62) \\
    \cellcolor{white} & ST-CAO & 35 & 0 & (0.39, 0.60) & 29 & 0 & (0.38, 0.59) \\
    \multirow{-4}{*}{\cellcolor{white} LR} & STv-CCTv-CAO & 36 & 0 & (0.44, 0.65) & 31 & 0 & (0.41, 0.62) \\
    \midrule
    \cellcolor{white} & STv-CTv & 42 & \bfseries 8 & \bfseries (0.56, 0.76) & 41 & \bfseries 9 & \bfseries (0.71, 0.88) \\
    \cellcolor{white} & STv-CCTv & 86 & \bfseries 41 & \bfseries (0.94, 1.00) & 80 & \bfseries 44 & \bfseries (0.96, 1.00) \\
    \cellcolor{white} & ST-CAO & 116 & \bfseries 70 & \bfseries (0.96, 1.00) & 114 & \bfseries 73 & \bfseries (0.96, 1.00) \\
    \multirow{-4}{*}{\cellcolor{white} RF} & STv-CCTv-CAO & 115 & \bfseries 72 & \bfseries (0.96, 1.00) & 116 & \bfseries 75 & \bfseries (0.96, 1.00) \\
    \midrule
    \cellcolor{white} & STv-CTv & 32 & 0 & (0.38, 0.61) & 22 & 0 & (0.40, 0.62) \\
    \cellcolor{white} & STv-CCTv & 39 & 1 & (0.45, 0.65) & 30 & 0 & (0.39, 0.60) \\
    \cellcolor{white} & ST-CAO & 42 & \bfseries 8 & \bfseries (0.62, 0.80) & 41 & \bfseries 12 & \bfseries (0.61, 0.79) \\
    \multirow{-4}{*}{\cellcolor{white} SVM} & STv-CCTv-CAO & 46 & \bfseries 9 & \bfseries (0.60, 0.79) & 41 & \bfseries 12 & \bfseries (0.68, 0.85) \\
    \bottomrule
    \end{tabular}
\end{table*}

\begin{table*}[htbp]
    \caption{DUD-E ROC enrichment at 0.01 FPR.}
    \centering
    \small
    \rowcolors{6}{}{lightgray}
    \sisetup{detect-weight=true,detect-inline-weight=math}
    \begin{tabular}{ l l S S c S S c }
    \toprule
     & & \multicolumn{3}{c}{Crystal Conformer} &
         \multicolumn{3}{c}{Generated Conformer} \\
    \cmidrule(lr){3-5} \cmidrule(lr){6-8}
    Model & Features &
    {\makecell{Median \\ $E_{0.01}$}} &
    {\makecell{Median \\ $\Delta E_{0.01}$}} &
    \makecell{Sign Test \\ 95\% CI} &
    {\makecell{Median \\ $E_{0.01}$}} &
    {\makecell{Median \\ $\Delta E_{0.01}$}} &
    \makecell{Sign Test \\ 95\% CI} \\
    \midrule
    \cellcolor{white} ROCS & TverskyCombo & 21 &  &  & 15 &  &  \\
    \midrule
    \cellcolor{white} & STv-CTv & 22 & 0 & (0.43, 0.66) & 15 & 0 & (0.36, 0.59) \\
    \cellcolor{white} & STv-CCTv & 25 & 1 & (0.46, 0.66) & 22 & 1 & (0.50, 0.70) \\
    \cellcolor{white} & ST-CAO & 23 & 0 & (0.44, 0.64) & 22 & 1 & (0.47, 0.68) \\
    \multirow{-4}{*}{\cellcolor{white} LR} & STv-CCTv-CAO & 26 & \bfseries 2 & \bfseries (0.53, 0.73) & 22 & \bfseries 3 & \bfseries (0.52, 0.72) \\
    \midrule
    \cellcolor{white} & STv-CTv & 26 & \bfseries 4 & \bfseries (0.56, 0.75) & 25 & \bfseries 5 & \bfseries (0.67, 0.84) \\
    \cellcolor{white} & STv-CCTv & 48 & \bfseries 22 & \bfseries (0.94, 1.00) & 44 & \bfseries 23 & \bfseries (0.92, 0.99) \\
    \cellcolor{white} & ST-CAO & 62 & \bfseries 36 & \bfseries (0.96, 1.00) & 62 & \bfseries 38 & \bfseries (0.96, 1.00) \\
    \multirow{-4}{*}{\cellcolor{white} RF} & STv-CCTv-CAO & 62 & \bfseries 36 & \bfseries (0.96, 1.00) & 62 & \bfseries 38 & \bfseries (0.96, 1.00) \\
    \midrule
    \cellcolor{white} & STv-CTv & 24 & 0 & (0.47, 0.68) & 16 & 0 & (0.43, 0.65) \\
    \cellcolor{white} & STv-CCTv & 24 & \bfseries 2 & \bfseries (0.55, 0.75) & 19 & 1 & (0.48, 0.68) \\
    \cellcolor{white} & ST-CAO & 29 & \bfseries 6 & \bfseries (0.74, 0.90) & 28 & \bfseries 8 & \bfseries (0.75, 0.90) \\
    \multirow{-4}{*}{\cellcolor{white} SVM} & STv-CCTv-CAO & 31 & \bfseries 7 & \bfseries (0.72, 0.88) & 30 & \bfseries 9 & \bfseries (0.73, 0.89) \\
    \bottomrule
    \end{tabular}
\end{table*}

\begin{table*}[htbp]
    \caption{DUD-E ROC enrichment at 0.02 FPR.}
    \centering
    \small
    \rowcolors{6}{}{lightgray}
    \sisetup{detect-weight=true,detect-inline-weight=math}
    \begin{tabular}{ l l S S c S S c }
    \toprule
     & & \multicolumn{3}{c}{Crystal Conformer} &
         \multicolumn{3}{c}{Generated Conformer} \\
    \cmidrule(lr){3-5} \cmidrule(lr){6-8}
    Model & Features &
    {\makecell{Median \\ $E_{0.02}$}} &
    {\makecell{Median \\ $\Delta E_{0.02}$}} &
    \makecell{Sign Test \\ 95\% CI} &
    {\makecell{Median \\ $E_{0.02}$}} &
    {\makecell{Median \\ $\Delta E_{0.02}$}} &
    \makecell{Sign Test \\ 95\% CI} \\
    \midrule
    \cellcolor{white} ROCS & TverskyCombo & 14 &  &  & 9 &  &  \\
    \midrule
    \cellcolor{white} & STv-CTv & 14 & 0 & (0.41, 0.63) & 10 & 0 & (0.45, 0.67) \\
    \cellcolor{white} & STv-CCTv & 15 & 0 & (0.45, 0.66) & 14 & 1 & (0.49, 0.69) \\
    \cellcolor{white} & ST-CAO & 15 & 1 & (0.48, 0.68) & 14 & \bfseries 1 & \bfseries (0.54, 0.74) \\
    \multirow{-4}{*}{\cellcolor{white} LR} & STv-CCTv-CAO & 17 & \bfseries 2 & \bfseries (0.57, 0.77) & 15 & \bfseries 3 & \bfseries (0.62, 0.80) \\
    \midrule
    \cellcolor{white} & STv-CTv & 15 & \bfseries 1 & \bfseries (0.50, 0.70) & 15 & \bfseries 2 & \bfseries (0.67, 0.84) \\
    \cellcolor{white} & STv-CCTv & 26 & \bfseries 10 & \bfseries (0.92, 0.99) & 25 & \bfseries 12 & \bfseries (0.91, 0.99) \\
    \cellcolor{white} & ST-CAO & 34 & \bfseries 18 & \bfseries (0.96, 1.00) & 33 & \bfseries 20 & \bfseries (0.96, 1.00) \\
    \multirow{-4}{*}{\cellcolor{white} RF} & STv-CCTv-CAO & 34 & \bfseries 17 & \bfseries (0.96, 1.00) & 33 & \bfseries 19 & \bfseries (0.94, 1.00) \\
    \midrule
    \cellcolor{white} & STv-CTv & 14 & 0 & (0.44, 0.65) & 11 & 0 & (0.49, 0.70) \\
    \cellcolor{white} & STv-CCTv & 15 & \bfseries 1 & \bfseries (0.55, 0.75) & 14 & \bfseries 1 & \bfseries (0.62, 0.81) \\
    \cellcolor{white} & ST-CAO & 19 & \bfseries 6 & \bfseries (0.80, 0.94) & 20 & \bfseries 6 & \bfseries (0.82, 0.95) \\
    \multirow{-4}{*}{\cellcolor{white} SVM} & STv-CCTv-CAO & 20 & \bfseries 6 & \bfseries (0.75, 0.90) & 20 & \bfseries 7 & \bfseries (0.79, 0.93) \\
    \bottomrule
    \end{tabular}
\end{table*}

\begin{table*}[htbp]
    \caption{DUD-E ROC enrichment at 0.05 FPR.}
    \centering
    \small
    \rowcolors{6}{}{lightgray}
    \sisetup{detect-weight=true,detect-inline-weight=math}
    \begin{tabular}{ l l S S c S S c }
    \toprule
     & & \multicolumn{3}{c}{Crystal Conformer} &
         \multicolumn{3}{c}{Generated Conformer} \\
    \cmidrule(lr){3-5} \cmidrule(lr){6-8}
    Model & Features &
    {\makecell{Median \\ $E_{0.05}$}} &
    {\makecell{Median \\ $\Delta E_{0.05}$}} &
    \makecell{Sign Test \\ 95\% CI} &
    {\makecell{Median \\ $E_{0.05}$}} &
    {\makecell{Median \\ $\Delta E_{0.05}$}} &
    \makecell{Sign Test \\ 95\% CI} \\
    \midrule
    \cellcolor{white} ROCS & TverskyCombo & 8 &  &  & 6 &  &  \\
    \midrule
    \cellcolor{white} & STv-CTv & 8 & \bfseries 0 & \bfseries (0.54, 0.75) & 6 & 0 & (0.47, 0.69) \\
    \cellcolor{white} & STv-CCTv & 8 & \bfseries 0 & \bfseries (0.52, 0.73) & 7 & \bfseries 1 & \bfseries (0.67, 0.85) \\
    \cellcolor{white} & ST-CAO & 9 & \bfseries 1 & \bfseries (0.66, 0.83) & 8 & \bfseries 2 & \bfseries (0.71, 0.87) \\
    \multirow{-4}{*}{\cellcolor{white} LR} & STv-CCTv-CAO & 10 & \bfseries 1 & \bfseries (0.78, 0.92) & 9 & \bfseries 2 & \bfseries (0.76, 0.91) \\
    \midrule
    \cellcolor{white} & STv-CTv & 8 & 0 & (0.49, 0.69) & 8 & \bfseries 1 & \bfseries (0.58, 0.77) \\
    \cellcolor{white} & STv-CCTv & 12 & \bfseries 4 & \bfseries (0.96, 1.00) & 11 & \bfseries 5 & \bfseries (0.94, 1.00) \\
    \cellcolor{white} & ST-CAO & 15 & \bfseries 7 & \bfseries (0.94, 1.00) & 14 & \bfseries 8 & \bfseries (0.92, 0.99) \\
    \multirow{-4}{*}{\cellcolor{white} RF} & STv-CCTv-CAO & 15 & \bfseries 7 & \bfseries (0.92, 0.99) & 14 & \bfseries 8 & \bfseries (0.94, 1.00) \\
    \midrule
    \cellcolor{white} & STv-CTv & 8 & \bfseries 0 & \bfseries (0.54, 0.75) & 7 & \bfseries 0 & \bfseries (0.64, 0.83) \\
    \cellcolor{white} & STv-CCTv & 9 & \bfseries 1 & \bfseries (0.70, 0.87) & 8 & \bfseries 1 & \bfseries (0.78, 0.93) \\
    \cellcolor{white} & ST-CAO & 11 & \bfseries 4 & \bfseries (0.84, 0.96) & 11 & \bfseries 4 & \bfseries (0.89, 0.98) \\
    \multirow{-4}{*}{\cellcolor{white} SVM} & STv-CCTv-CAO & 12 & \bfseries 4 & \bfseries (0.87, 0.97) & 12 & \bfseries 4 & \bfseries (0.89, 0.98) \\
    \bottomrule
    \end{tabular}
\end{table*}

\subsubsection{MUV}

\begin{table*}[htbp]
    \caption{MUV ROC enrichment at 0.005 FPR.}
    \centering
    \small
    \rowcolors{2}{lightgray}{}
    \sisetup{detect-weight=true,detect-inline-weight=math}
    \begin{tabular}{ l l S S c }
    \toprule
    Model & Features &
    {\makecell{Median \\ $E_{0.005}$}} &
    {\makecell{Median \\ $\Delta E_{0.005}$}} &
    \makecell{Sign Test \\ 95\% CI} \\
    \midrule
    \cellcolor{white} ROCS & TverskyCombo & 0 &  &  \\
    \midrule
    \cellcolor{white} & STv-CTv & 0 & 0 & (0.40, 0.58) \\
    \cellcolor{white} & STv-CCTv & 0 & \bfseries \color{red} 0 & \bfseries \color{red} (0.30, 0.45) \\
    \cellcolor{white} & ST-CAO & 0 & 0 & (0.38, 0.52) \\
    \multirow{-4}{*}{\cellcolor{white} LR} & STv-CCTv-CAO & 0 & \bfseries \color{red} 0 & \bfseries \color{red} (0.34, 0.48) \\
    \midrule
    \cellcolor{white} & STv-CTv & 0 & 0 & (0.44, 0.57) \\
    \cellcolor{white} & STv-CCTv & 8 & \bfseries 0 & \bfseries (0.58, 0.70) \\
    \cellcolor{white} & ST-CAO & 10 & \bfseries 6 & \bfseries (0.67, 0.77) \\
    \multirow{-4}{*}{\cellcolor{white} RF} & STv-CCTv-CAO & 10 & \bfseries 5 & \bfseries (0.67, 0.77) \\
    \midrule
    \cellcolor{white} & STv-CTv & 0 & \bfseries \color{red} 0 & \bfseries \color{red} (0.31, 0.47) \\
    \cellcolor{white} & STv-CCTv & 0 & \bfseries \color{red} 0 & \bfseries \color{red} (0.25, 0.39) \\
    \cellcolor{white} & ST-CAO & 0 & 0 & (0.37, 0.51) \\
    \multirow{-4}{*}{\cellcolor{white} SVM} & STv-CCTv-CAO & 0 & \bfseries \color{red} 0 & \bfseries \color{red} (0.33, 0.46) \\
    \bottomrule
    \end{tabular}
\end{table*}

\begin{table*}[htbp]
    \caption{MUV ROC enrichment at 0.01 FPR.}
    \centering
    \small
    \rowcolors{2}{lightgray}{}
    \sisetup{detect-weight=true,detect-inline-weight=math}
    \begin{tabular}{ l l S S c }
    \toprule
    Model & Features &
    {\makecell{Median \\ $E_{0.01}$}} &
    {\makecell{Median \\ $\Delta E_{0.01}$}} &
    \makecell{Sign Test \\ 95\% CI} \\
    \midrule
    \cellcolor{white} ROCS & TverskyCombo & 4 &  &  \\
    \midrule
    \cellcolor{white} & STv-CTv & 4 & 0 & (0.40, 0.56) \\
    \cellcolor{white} & STv-CCTv & 0 & \bfseries \color{red} 0 & \bfseries \color{red} (0.33, 0.47) \\
    \cellcolor{white} & ST-CAO & 4 & 0 & (0.44, 0.57) \\
    \multirow{-4}{*}{\cellcolor{white} LR} & STv-CCTv-CAO & 0 & 0 & (0.39, 0.52) \\
    \midrule
    \cellcolor{white} & STv-CTv & 1 & 0 & (0.44, 0.56) \\
    \cellcolor{white} & STv-CCTv & 5 & \bfseries 0 & \bfseries (0.57, 0.68) \\
    \cellcolor{white} & ST-CAO & 7 & \bfseries 4 & \bfseries (0.63, 0.73) \\
    \multirow{-4}{*}{\cellcolor{white} RF} & STv-CCTv-CAO & 8 & \bfseries 4 & \bfseries (0.66, 0.76) \\
    \midrule
    \cellcolor{white} & STv-CTv & 0 & \bfseries \color{red} 0 & \bfseries \color{red} (0.32, 0.46) \\
    \cellcolor{white} & STv-CCTv & 0 & \bfseries \color{red} 0 & \bfseries \color{red} (0.27, 0.40) \\
    \cellcolor{white} & ST-CAO & 4 & 0 & (0.44, 0.56) \\
    \multirow{-4}{*}{\cellcolor{white} SVM} & STv-CCTv-CAO & 2 & 0 & (0.41, 0.53) \\
    \bottomrule
    \end{tabular}
\end{table*}

\begin{table*}[htbp]
    \caption{MUV ROC enrichment at 0.02 FPR.}
    \centering
    \small
    \rowcolors{2}{lightgray}{}
    \sisetup{detect-weight=true,detect-inline-weight=math}
    \begin{tabular}{ l l S S c }
    \toprule
    Model & Features &
    {\makecell{Median \\ $E_{0.02}$}} &
    {\makecell{Median \\ $\Delta E_{0.02}$}} &
    \makecell{Sign Test \\ 95\% CI} \\
    \midrule
    \cellcolor{white} ROCS & TverskyCombo & 2 &  &  \\
    \midrule
    \cellcolor{white} & STv-CTv & 2 & 0 & (0.41, 0.55) \\
    \cellcolor{white} & STv-CCTv & 2 & \bfseries \color{red} 0 & \bfseries \color{red} (0.38, 0.50) \\
    \cellcolor{white} & ST-CAO & 2 & 0 & (0.46, 0.58) \\
    \multirow{-4}{*}{\cellcolor{white} LR} & STv-CCTv-CAO & 2 & 0 & (0.45, 0.57) \\
    \midrule
    \cellcolor{white} & STv-CTv & 2 & 0 & (0.47, 0.58) \\
    \cellcolor{white} & STv-CCTv & 4 & \bfseries 0 & \bfseries (0.54, 0.65) \\
    \cellcolor{white} & ST-CAO & 4 & \bfseries 2 & \bfseries (0.62, 0.72) \\
    \multirow{-4}{*}{\cellcolor{white} RF} & STv-CCTv-CAO & 5 & \bfseries 2 & \bfseries (0.65, 0.75) \\
    \midrule
    \cellcolor{white} & STv-CTv & 2 & \bfseries \color{red} 0 & \bfseries \color{red} (0.30, 0.42) \\
    \cellcolor{white} & STv-CCTv & 2 & \bfseries \color{red} 0 & \bfseries \color{red} (0.31, 0.43) \\
    \cellcolor{white} & ST-CAO & 2 & \bfseries 0 & \bfseries (0.53, 0.64) \\
    \multirow{-4}{*}{\cellcolor{white} SVM} & STv-CCTv-CAO & 2 & 0 & (0.47, 0.58) \\
    \bottomrule
    \end{tabular}
\end{table*}

\begin{table*}[htbp]
    \caption{MUV ROC enrichment at 0.05 FPR.}
    \centering
    \small
    \rowcolors{2}{lightgray}{}
    \sisetup{detect-weight=true,detect-inline-weight=math}
    \begin{tabular}{ l l S S c }
    \toprule
    Model & Features &
    {\makecell{Median \\ $E_{0.05}$}} &
    {\makecell{Median \\ $\Delta E_{0.05}$}} &
    \makecell{Sign Test \\ 95\% CI} \\
    \midrule
    \cellcolor{white} ROCS & TverskyCombo & 2 &  &  \\
    \midrule
    \cellcolor{white} & STv-CTv & 2 & 0 & (0.47, 0.60) \\
    \cellcolor{white} & STv-CCTv & 2 & 0 & (0.44, 0.55) \\
    \cellcolor{white} & ST-CAO & 2 & \bfseries 0 & \bfseries (0.53, 0.64) \\
    \multirow{-4}{*}{\cellcolor{white} LR} & STv-CCTv-CAO & 2 & \bfseries 0 & \bfseries (0.52, 0.63) \\
    \midrule
    \cellcolor{white} & STv-CTv & 2 & 0 & (0.45, 0.55) \\
    \cellcolor{white} & STv-CCTv & 2 & \bfseries 0 & \bfseries (0.50, 0.60) \\
    \cellcolor{white} & ST-CAO & 3 & \bfseries 1 & \bfseries (0.58, 0.68) \\
    \multirow{-4}{*}{\cellcolor{white} RF} & STv-CCTv-CAO & 3 & \bfseries 1 & \bfseries (0.61, 0.70) \\
    \midrule
    \cellcolor{white} & STv-CTv & 2 & 0 & (0.40, 0.52) \\
    \cellcolor{white} & STv-CCTv & 2 & \bfseries \color{red} 0 & \bfseries \color{red} (0.39, 0.50) \\
    \cellcolor{white} & ST-CAO & 3 & \bfseries 1 & \bfseries (0.56, 0.67) \\
    \multirow{-4}{*}{\cellcolor{white} SVM} & STv-CCTv-CAO & 2 & \bfseries 1 & \bfseries (0.54, 0.64) \\
    \bottomrule
    \end{tabular}
\end{table*}

\subsubsection{ChEMBL}

\begin{table*}[htbp]
    \caption{ChEMBL ROC enrichment at 0.005 FPR.}
    \centering
    \small
    \rowcolors{2}{lightgray}{}
    \sisetup{detect-weight=true,detect-inline-weight=math}
    \begin{tabular}{ l l S S c }
    \toprule
    Model & Features &
    {\makecell{Median \\ $E_{0.005}$}} &
    {\makecell{Median \\ $\Delta E_{0.005}$}} &
    \makecell{Sign Test \\ 95\% CI} \\
    \midrule
    \cellcolor{white} ROCS & TverskyCombo & 21 &  &  \\
    \midrule
    \cellcolor{white} & STv-CTv & 24 & \bfseries 0 & \bfseries (0.52, 0.55) \\
    \cellcolor{white} & STv-CCTv & 23 & \bfseries 0 & \bfseries (0.52, 0.55) \\
    \cellcolor{white} & ST-CAO & 21 & 0 & (0.48, 0.52) \\
    \multirow{-4}{*}{\cellcolor{white} LR} & STv-CCTv-CAO & 22 & \bfseries 0 & \bfseries (0.51, 0.54) \\
    \midrule
    \cellcolor{white} & STv-CTv & 24 & \bfseries 0 & \bfseries (0.51, 0.54) \\
    \cellcolor{white} & STv-CCTv & 45 & \bfseries 18 & \bfseries (0.83, 0.85) \\
    \cellcolor{white} & ST-CAO & 60 & \bfseries 32 & \bfseries (0.90, 0.92) \\
    \multirow{-4}{*}{\cellcolor{white} RF} & STv-CCTv-CAO & 65 & \bfseries 35 & \bfseries (0.91, 0.93) \\
    \midrule
    \cellcolor{white} & STv-CTv & 23 & 0 & (0.48, 0.52) \\
    \cellcolor{white} & STv-CCTv & 22 & 0 & (0.48, 0.52) \\
    \cellcolor{white} & ST-CAO & 21 & 0 & (0.49, 0.53) \\
    \multirow{-4}{*}{\cellcolor{white} SVM} & STv-CCTv-CAO & 25 & \bfseries 3 & \bfseries (0.56, 0.59) \\
    \bottomrule
    \end{tabular}
\end{table*}

\begin{table*}[htbp]
    \caption{ChEMBL ROC enrichment at 0.01 FPR.}
    \centering
    \small
    \rowcolors{2}{lightgray}{}
    \sisetup{detect-weight=true,detect-inline-weight=math}
    \begin{tabular}{ l l S S c }
    \toprule
    Model & Features &
    {\makecell{Median \\ $E_{0.01}$}} &
    {\makecell{Median \\ $\Delta E_{0.01}$}} &
    \makecell{Sign Test \\ 95\% CI} \\
    \midrule
    \cellcolor{white} ROCS & TverskyCombo & 13 &  &  \\
    \midrule
    \cellcolor{white} & STv-CTv & 16 & \bfseries 0 & \bfseries (0.56, 0.60) \\
    \cellcolor{white} & STv-CCTv & 16 & \bfseries 1 & \bfseries (0.59, 0.62) \\
    \cellcolor{white} & ST-CAO & 16 & \bfseries 1 & \bfseries (0.56, 0.59) \\
    \multirow{-4}{*}{\cellcolor{white} LR} & STv-CCTv-CAO & 18 & \bfseries 4 & \bfseries (0.65, 0.68) \\
    \midrule
    \cellcolor{white} & STv-CTv & 15 & \bfseries 0 & \bfseries (0.51, 0.54) \\
    \cellcolor{white} & STv-CCTv & 28 & \bfseries 11 & \bfseries (0.84, 0.86) \\
    \cellcolor{white} & ST-CAO & 36 & \bfseries 19 & \bfseries (0.91, 0.93) \\
    \multirow{-4}{*}{\cellcolor{white} RF} & STv-CCTv-CAO & 38 & \bfseries 20 & \bfseries (0.92, 0.94) \\
    \midrule
    \cellcolor{white} & STv-CTv & 16 & \bfseries 0 & \bfseries (0.54, 0.57) \\
    \cellcolor{white} & STv-CCTv & 16 & \bfseries 1 & \bfseries (0.56, 0.59) \\
    \cellcolor{white} & ST-CAO & 17 & \bfseries 2 & \bfseries (0.60, 0.63) \\
    \multirow{-4}{*}{\cellcolor{white} SVM} & STv-CCTv-CAO & 19 & \bfseries 4 & \bfseries (0.69, 0.72) \\
    \bottomrule
    \end{tabular}
\end{table*}

\begin{table*}[htbp]
    \caption{ChEMBL ROC enrichment at 0.02 FPR.}
    \centering
    \small
    \rowcolors{2}{lightgray}{}
    \sisetup{detect-weight=true,detect-inline-weight=math}
    \begin{tabular}{ l l S S c }
    \toprule
    Model & Features &
    {\makecell{Median \\ $E_{0.02}$}} &
    {\makecell{Median \\ $\Delta E_{0.02}$}} &
    \makecell{Sign Test \\ 95\% CI} \\
    \midrule
    \cellcolor{white} ROCS & TverskyCombo & 9 &  &  \\
    \midrule
    \cellcolor{white} & STv-CTv & 10 & \bfseries 0 & \bfseries (0.58, 0.62) \\
    \cellcolor{white} & STv-CCTv & 11 & \bfseries 1 & \bfseries (0.65, 0.68) \\
    \cellcolor{white} & ST-CAO & 11 & \bfseries 2 & \bfseries (0.63, 0.66) \\
    \multirow{-4}{*}{\cellcolor{white} LR} & STv-CCTv-CAO & 14 & \bfseries 4 & \bfseries (0.79, 0.82) \\
    \midrule
    \cellcolor{white} & STv-CTv & 10 & 0 & (0.49, 0.53) \\
    \cellcolor{white} & STv-CCTv & 17 & \bfseries 6 & \bfseries (0.85, 0.87) \\
    \cellcolor{white} & ST-CAO & 22 & \bfseries 10 & \bfseries (0.93, 0.94) \\
    \multirow{-4}{*}{\cellcolor{white} RF} & STv-CCTv-CAO & 23 & \bfseries 11 & \bfseries (0.94, 0.95) \\
    \midrule
    \cellcolor{white} & STv-CTv & 11 & \bfseries 0 & \bfseries (0.59, 0.63) \\
    \cellcolor{white} & STv-CCTv & 11 & \bfseries 2 & \bfseries (0.66, 0.69) \\
    \cellcolor{white} & ST-CAO & 13 & \bfseries 3 & \bfseries (0.71, 0.74) \\
    \multirow{-4}{*}{\cellcolor{white} SVM} & STv-CCTv-CAO & 14 & \bfseries 4 & \bfseries (0.80, 0.82) \\
    \bottomrule
    \end{tabular}
\end{table*}

\begin{table*}[htbp]
    \caption{ChEMBL ROC enrichment at 0.05 FPR.}
    \centering
    \small
    \rowcolors{2}{lightgray}{}
    \sisetup{detect-weight=true,detect-inline-weight=math}
    \begin{tabular}{ l l S S c }
    \toprule
    Model & Features &
    {\makecell{Median \\ $E_{0.05}$}} &
    {\makecell{Median \\ $\Delta E_{0.05}$}} &
    \makecell{Sign Test \\ 95\% CI} \\
    \midrule
    \cellcolor{white} ROCS & TverskyCombo & 5 &  &  \\
    \midrule
    \cellcolor{white} & STv-CTv & 6 & \bfseries 0 & \bfseries (0.62, 0.65) \\
    \cellcolor{white} & STv-CCTv & 7 & \bfseries 1 & \bfseries (0.73, 0.76) \\
    \cellcolor{white} & ST-CAO & 7 & \bfseries 1 & \bfseries (0.69, 0.72) \\
    \multirow{-4}{*}{\cellcolor{white} LR} & STv-CCTv-CAO & 9 & \bfseries 3 & \bfseries (0.89, 0.91) \\
    \midrule
    \cellcolor{white} & STv-CTv & 6 & 0 & (0.50, 0.53) \\
    \cellcolor{white} & STv-CCTv & 9 & \bfseries 3 & \bfseries (0.84, 0.86) \\
    \cellcolor{white} & ST-CAO & 11 & \bfseries 4 & \bfseries (0.92, 0.94) \\
    \multirow{-4}{*}{\cellcolor{white} RF} & STv-CCTv-CAO & 11 & \bfseries 5 & \bfseries (0.94, 0.95) \\
    \midrule
    \cellcolor{white} & STv-CTv & 6 & \bfseries 0 & \bfseries (0.68, 0.71) \\
    \cellcolor{white} & STv-CCTv & 7 & \bfseries 1 & \bfseries (0.78, 0.81) \\
    \cellcolor{white} & ST-CAO & 8 & \bfseries 3 & \bfseries (0.80, 0.83) \\
    \multirow{-4}{*}{\cellcolor{white} SVM} & STv-CCTv-CAO & 9 & \bfseries 3 & \bfseries (0.86, 0.88) \\
    \bottomrule
    \end{tabular}
\end{table*}

\putbib
\end{bibunit}

\end{document}